\documentclass[letterpaper]{article} 
\usepackage{aaai2026}  
\usepackage{times}  
\usepackage{helvet}  
\usepackage{courier}  
\usepackage[hyphens]{url}  
\usepackage{graphicx} 
\usepackage{natbib}  
\usepackage{caption} 
\usepackage{algorithm}
\usepackage{algorithmic}

\usepackage{newfloat}
\usepackage{listings}
\DeclareCaptionStyle{ruled}{labelfont=normalfont,labelsep=colon,strut=off} 
\floatstyle{ruled}
\newfloat{listing}{tb}{lst}{}
\floatname{listing}{Listing}
\usepackage{booktabs}
\usepackage{xcolor,colortbl}
\usepackage{subcaption}
\usepackage{amsmath}
\usepackage{csquotes}

\newcommand{\answerYes}[1]{\textcolor{blue}{#1}} 
\newcommand{\answerNo}[1]{\textcolor{teal}{#1}} 
\newcommand{\answerNA}[1]{\textcolor{gray}{#1}}

\title{Politics of Questions in News: A Mixed-Methods Study of Interrogative Stances as Markers of Voice and Power}
\author {
    Victor Bros\textsuperscript{\rm 1, \rm 2},
    Matilde Barbini\textsuperscript{\rm 2},
    Patrick Gerard\textsuperscript{\rm 3},
    Daniel Gatica-Perez\textsuperscript{\rm 1, \rm 2}
}
\affiliations {
    \textsuperscript{\rm 1}Idiap Research Institute, Switzerland\\
    \textsuperscript{\rm 2}EPFL, Switzerland\\
    \textsuperscript{\rm 3}University of Southern California, USA\\
    victor.bros@idiap.ch, matilde.barbini@epfl.ch, pgerard@isi.edu, gatica@idiap.ch
}

\begin{document}

\maketitle

\begin{abstract}
Interrogatives in news discourse have been examined in linguistics and conversation analysis, but mostly in broadcast interviews and relatively small, often English-language corpora, while large-scale computational studies of news rarely distinguish interrogatives from declaratives or differentiate their functions. This paper brings these strands together through a mixed-methods study of the \enquote{Politics of Questions} in contemporary French-language digital news.
Using over one million articles published between January 2023 and June 2024, we automatically detect interrogative stances, approximate their functional types, and locate textual answers when present, linking these quantitative measures to a qualitatively annotated subcorpus grounded in semantic and pragmatic theories of questions.
Interrogatives are sparse but systematically patterned: they mainly introduce or organize issues, with most remaining cases being information-seeking or echo-like, while explicitly leading or tag questions are rare. Although their density and mix vary across outlets and topics, our heuristic suggests that questions are overwhelmingly taken up within the same article and usually linked to a subsequent answer-like span, most often in the journalist’s narrative voice and less often through quoted speech. Interrogative contexts are densely populated with named individuals, organizations, and places, whereas publics and broad social groups are mentioned much less frequently, suggesting that interrogative discourse tends to foreground already prominent actors and places and thus exhibits strong personalization. We show how interrogative stance, textual uptake, and voice can be operationalized at corpus scale, and argue that combining computational methods with pragmatic and sociological perspectives can help account for how questioning practices structure contemporary news discourse.

\end{abstract}

\section{Introduction}
\label{sec:intro}

The digital transition has reshaped how news is produced, financed, and encountered. Audiences increasingly consume news online and via platforms, while legacy outlets face intensified competition for attention and pressure on business models~\cite{hindman_internet_2018,FOEG_Yearbook_2024}. In many countries this has contributed to \enquote{news deserts}, where local reporting capacity erodes and communities lose access to robust, professional journalism, with documented democratic harms~\cite{abernathy_news_2023,vogler_investigating_2023,shaker_dead_2014,barclay_local_2025}. Against this backdrop, questions about who and what is made visible in news, whose problems are problematized, and how publics are addressed acquire renewed democratic significance~\cite{pickard_democracy_2020}.

These democratic roles are enacted not only through topic selection and framing, but also through interactional practices such as questioning and challenging. In broadcast settings, conversation-analytic work has shown how question design ranges from deferential information-seeking to highly adversarial, accountability-pressuring formats~\cite{greatbatch_turn-taking_1988,clayman_news_2002,heritage_epistemics_2012}. Linguistics and pragmatics emphasize that interrogatives are context-dependent acts that can request information, highlight issues, embed presuppositions, or function rhetorically~\cite{athanasiadou_modes_1991,ilie_what_1994,van_rooy_questioning_2003,ciardelli_questions_2016}. Most of this work relies on close analysis of relatively small, often English-language corpora. Computational studies of rhetorical questions and stance in online debate~\cite{ranganath_identifying_2016,oraby_are_2017} show that some pragmatic distinctions can be modeled automatically, but remain limited in genre and scale. Large-scale NLP studies of news, in turn, focus mainly on topics, sentiment, or frames and usually treat all sentences alike, without distinguishing interrogatives from declaratives or separating different interrogative functions.

French-language news provides a particularly interesting setting for bridging these strands. It spans multiple media systems and political contexts, including France, Switzerland, Belgium, francophone Canada, Senegal, and pan-African outlets, with distinct histories of press freedom and journalistic cultures~\cite{hallin_comparing_2004,straubhaar_cultural_2021}. Yet French functions as a shared linguistic resource, enabling partly common repertoires of questioning and address while leaving room for local variation in how power and publics are constructed. This combination makes francophone news a useful testbed for our approach: it moves beyond the Anglophone focus of much prior work while keeping a single language and covering both Global North and Global South media systems. We bring a mixed-methods, computational lens to what we call the \emph{Politics of Questions} in this fragmented but interconnected media space. Using more than 1.2 million French-language digital news articles (January 2023-June 2024), we automatically identify interrogative stances in context, approximate their functional types, locate textual answers when they are present, and characterize the actors and places that populate questions and answers. These quantitative components are linked to a qualitatively annotated subcorpus, grounded in semantic and pragmatic theories of interrogatives, which we use for both evaluation and discourse analysis.

Focusing on \emph{interrogative stances} allows us to make the politics of visibility and accountability empirically tractable. Questions are a particularly salient subset of journalistic moves: they mark issues as open, allocate epistemic authority, and can be detected and typed from surface cues in ways that most other interactional practices cannot. At the same time, they are normatively charged, because asking or not asking, and how a question is framed, can shape which actors are foregrounded, which problems are rendered discussable, and how accountability is textually staged.

This leads to the following research questions:

\noindent\textbf{RQ1}: How prevalent are interrogative stances in French-language news, and how do their types vary across outlets, countries, and topics?

\noindent\textbf{RQ2}: When journalists pose interrogative stances, to what extent are those questions textually answered, and how often do answers involve quoted external voices versus self-answers by the journalist?

\noindent\textbf{RQ3}: Which kinds of actors and social groups are most often foregrounded in interrogative contexts, and how frequently do such contexts include quoted external voices rather than only journalistic narration?

\noindent\textbf{RQ4}: How can a triangulated methodology, combining quantitative and qualitative analyses, enhance the interpretation and validity of the quantitative measures?

\noindent\textbf{Contributions.} Our contributions are threefold: (i) conceptually, we extend semantic and pragmatic work on questions to written news by operationalizing \emph{interrogative stances}, answerhood, and dialogicity at corpus scale; (ii) empirically, we provide, to our knowledge, the first large-scale, cross-regional description of interrogative practices in French-language digital news, showing how interrogatives structure coverage, how often they are resolved in-text, and which actors and places they center; and (iii) methodologically, we propose and validate a triangulated pipeline in which large language models and fine-tuned neural classifiers are combined with qualitative annotation to yield interpretable proxy measures relevant to debates on gatekeeping, voice, and agenda-setting.

\section{Related Work}
\label{sec:related}

\paragraph{Theories of questions: semantics and pragmatics}

Linguistics and philosophy of language treat questions not just as syntactic configurations but as discourse moves whose interpretation depends on context. Formal semantics represents questions as abstract objects that partition the space of possible worlds and determine which answers resolve an issue~\cite{stokhof_jeroen_nodate,ciardelli_inquisitive_2018}, while contextual models such as Questions Under Discussion and Table-based accounts~\cite{farkas_reacting_2010,ginzburg_interactive_2015} embed this view in a dynamic picture of interaction where questions guide relevance and shape how common ground is updated. Pragmatic studies emphasize that interrogatives must be classified according to speaker intentions and social situations: they can request information, highlight issues, regulate knowledge display, or indirectly request action~\cite{hudson_meaning_1975,athanasiadou_modes_1991}. Classic distinctions between information-seeking, rhetorical, leading, and echo questions, and between more or less \enquote{conducive} designs, show how presuppositions and turn design can steer respondents toward particular answers or implicatures~\cite{van_rooy_questioning_2003,ilie_what_1994,frank_you_1990,braun_implicating_2011}. These insights underpin our use of \emph{interrogative stance} as a contextual property of utterances and motivate the stance types we distinguish in the empirical analysis.

\paragraph{Questions, accountability, framing, and gatekeeping in news discourse}

Beyond their semantic content, questions enact social relations and allocate accountability. Athanasiadou~\cite{athanasiadou_modes_1991} notes that questioning carries a \enquote{dominance function} tied to roles: the same interrogative can be heard as a neutral request among peers or as a command in hierarchical settings. Stivers and Rossano~\cite{stivers_mobilizing_2010} show that interrogative turns mobilize responses with varying strength depending on action, design, and sequential position. In news interviews, Greatbatch~\cite{greatbatch_turn-taking_1988} and Clayman and Heritage~\cite{clayman_news_2002} document how journalists control topical progression and use question design to negotiate norms of neutrality and adversarialness. Diachronic work on U.S. presidential press conferences finds a rise in more direct and negative interrogatives, interpreted as evidence of increasing pressure on presidents to respond in particular ways~\cite{clayman_pressuring_2023}. Media sociology and communication research add a broader perspective on how news institutions filter and present public issues. Classic gatekeeping work shows how newsroom routines and professional values prioritize certain topics, locations, and source types, leading to the overrepresentation of political and economic elites~\cite{gans_deciding_2004,shoemaker_gatekeeping_2009}. Framing theory conceptualizes how media selectively emphasize certain aspects of reality to promote particular problem definitions, causal stories, or remedies~\cite{entman_framing_1993}. Our focus on interrogative stances, their answerability, and their actor populations connects these interactional and sociological perspectives by asking how questions help to distribute accountability and voice in news discourse.

\paragraph{Computational approaches to questions and news content}

Computational work has begun to model pragmatic functions of questions, but mostly in social media and with coarse distinctions. Ranganath et al.~\cite{ranganath_identifying_2016} and Oraby et al.~\cite{oraby_are_2017} compile corpora of rhetorical questions from online debate forums and Twitter and train classifiers that distinguish rhetorical from non-rhetorical and sarcastic from non-sarcastic questions with solid performance. These studies demonstrate that some pragmatic categories can be learned automatically, but focus on English user-generated dialogue and do not consider how questions are taken up or answered in subsequent discourse.

In journalism and communication studies, most computational analyses of text have focused on topics, frames, and temporal dynamics rather than interrogatives. Large-scale frame analyses apply supervised or semi-supervised models to millions of news articles to trace how frames fluctuate with events and evolve over longer periods~\cite{kwak_systematic_2020,guo_framing_2011,dehler-holland_topic_2021}. Other work combines topic modeling with time-series methods to study intermedia agenda-setting, for example in coverage of vaccine scandals or policy crises on different platforms~\cite{shi_examining_2023}. Recent studies use topic models and named-entity recognition to assess how well COVID-19 coverage fulfills normative functions such as informing, monitoring, and providing a platform for debate~\cite{nguyen_exploring_2024}. Across this literature, sentences are typically treated uniformly for modeling topics, frames, or sentiment, and questions are not analyzed as distinct interactional moves. Moreover, even when headlines or rhetorical devices are examined qualitatively, there is little systematic attention to the micro-pragmatics of questioning, such as conduciveness, presupposition, or dialogicity, across large corpora. Our work complements these approaches by foregrounding interrogative acts themselves and by linking question design, answerhood, and actor mentions to agenda-setting and gatekeeping.

\paragraph{Mixed-methods and triangulated computational media research}

A parallel methodological literature argues for combining computational text analysis with qualitative inquiry. Triangulation is understood not only as cross-validation but as bringing multiple kinds of evidence and perspectives to bear on the same phenomenon~\cite{olsen_triangulation_nodate}. Nelson~\cite{nelson_computational_2020} formulates this as \enquote{computational grounded theory}, in which unsupervised or weakly supervised methods surface patterns, qualitative analysis refines concepts, and further computational analyses test their scope.
Large language models are increasingly integrated into such pipelines as annotation or coding tools. Gilardi et al.~\cite{gilardi_chatgpt_2023} compare ChatGPT to crowdworkers and trained annotators on multiple text-coding tasks and show that zero-shot LLM classifications can rival crowdworkers in accuracy and intercoder agreement at much lower cost, though human oversight remains essential. We build on this line of work by using an LLM to generate high-confidence pseudo-labels of interrogative stance, distilling these into fine-tuned neural classifiers, and then interpreting their outputs through a qualitatively annotated subcorpus. This triangulated design allows us to operationalize pragmatic concepts such as interrogative stance, answerhood, and voice at scale while retaining a close connection to discourse-level analysis.

\section{Methodology}
\label{sec:method}

\begin{figure*}[t]
    \centering
    \includegraphics[width=\textwidth]{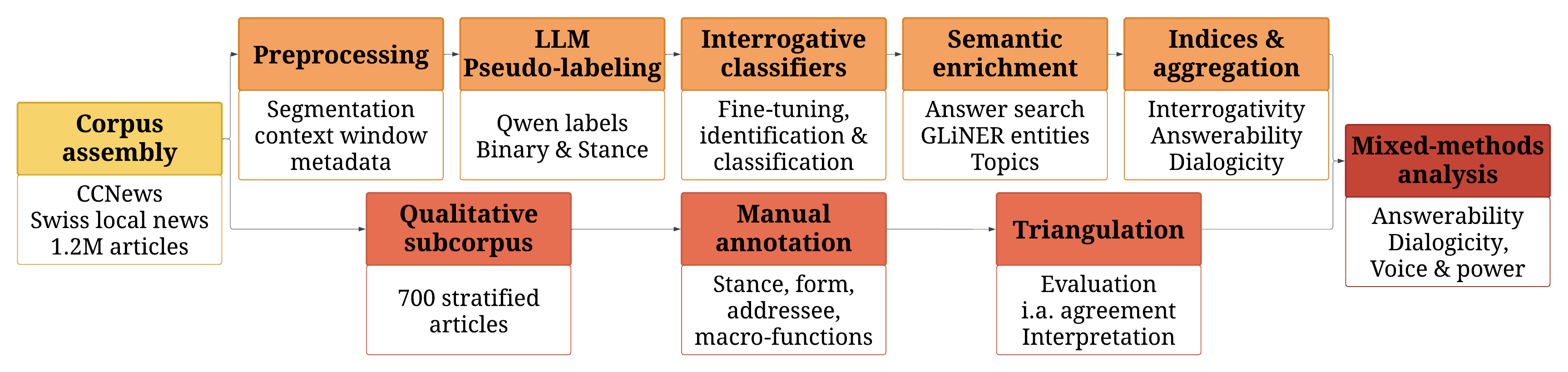}
    \caption{\textbf{Overview of the mixed-methods pipeline for interrogative stance analysis.} Starting from a corpus of 1.2M French-language news articles from 24 outlets (2023-2024), the upper branch applies a teacher-student NLP pipeline to detect interrogative stances and derive measures of interrogativity, answerability, dialogicity, and actor/addressivity profiles. In parallel, a stratified qualitative subcorpus is manually annotated for stance, form, addressee, answerhood, and macro-functions. The two branches are then triangulated to evaluate the models and interpret corpus-level patterns of prevalence, answerability, voice, and power.}
    \label{fig:method-overview}
\end{figure*}

\subsection{Data}

We assemble a French-language news corpus from two complementary sources. The first component is derived from CCNews, a large-scale news collection built from Common Crawl~\cite{CCNewsHamborg}. Starting from the most active French-language CCNews domains in our period (top 40 by article volume), we manually retained outlets that (i) publish in French, (ii) remain active in both 2023 and 2024, (iii) can be identified as established news organizations via their domains and websites, and (iv) yield sufficiently clean article extractions, excluding domains with substantial scraping noise. To avoid overconcentration in a single national market, we also prioritized coverage across distinct francophone regions. This yields just over 1.2 million articles spanning global, national, and regional coverage from French-speaking Europe, America, and Africa.

Hyper-local actors and issues are underrepresented in this CCNews-derived subset. To better cover local journalism, we integrate a second open dataset of Swiss local news~\cite{Bros_Gatica-Perez_2025}, which provides high-quality reporting from cantonal and hyper-local outlets in Romandy. From this collection we include all French-language articles published between 2023 and mid-2024. Combined, the final corpus comprises 24 outlets. For each outlet we manually code an \emph{editorial scale} (hyper-local, regional, national, transnational, thematic) and a primary \emph{country or region}, used in later analyses of cross-country and cross-scale variation. This corpus should therefore be read as a curated comparative sample rather than a statistically representative census of French-language digital news. All country- and scale-level aggregates should therefore be read as descriptive comparisons within this curated sample rather than representative estimates of francophone news systems, because outlet volumes are highly unequal, larger outlets contribute more to article-level group means. Full outlet lists and counts are provided in Appendix~\ref{app:data}.

\paragraph{Ethical considerations}

All data come from publicly accessible news websites or established web corpora. All external models and corpora we use are under permissive research licenses. We analyze text for scientific purposes only, report results in aggregate primarily, use a small number of short, translated and lightly paraphrased excerpts, and do not attempt to deanonymize individuals beyond what is already present in published news. We do not redistribute full-text articles. Any shared data will be limited to derived annotations and metadata, subject to the terms of use of the original sources.

\subsection{Definition of interrogative stance}
\label{subsec:definition_interrogative}

Following semantic and pragmatic work on questions~\cite{hudson_meaning_1975,ciardelli_questions_2016,ginzburg_interactive_2015,athanasiadou_modes_1991}, we treat \emph{interrogative stance} as a contextual property of utterances rather than a purely syntactic category. At the sentence level, a journalistic sentence expresses an interrogative stance if it does at least one of the following:

\begin{enumerate}
    \item Explicitly presents a question or problem to be answered (direct questions with \enquote{?} or indirect questions).
    \item Implicitly raises an issue or unknown whose resolution is projected in the surrounding discourse.
    \item Highlights that some aspect of the situation remains open, unresolved, or unanswered, thereby framing it as an issue rather than a settled fact.
\end{enumerate}

Sentences that merely mention a \enquote{question} or \enquote{issue} as a topic, without presenting a concrete unknown or projecting its resolution, are treated as non-interrogative. Likewise, purely evaluative statements of uncertainty are not classified as interrogative unless they are clearly framed as an open question to be addressed.

Building on pragmatic typologies~\cite{athanasiadou_modes_1991,ilie_what_1994,frank_you_1990,van_rooy_questioning_2003}, we distinguish six interrogative stance types:

\begin{itemize}
    \item \textbf{Framing-procedural}: interrogative or issue-raising sentences whose primary role is to structure the article (introduce a problem, announce a section) or to mark that a question remains open, rather than to request an answer from a specific interlocutor.
    \item \textbf{Information-seeking}: sentences that express a genuine request for factual information or clarification; the writer or quoted speaker is presented as lacking the information and expecting an answer.
    \item \textbf{Rhetorical}: formally interrogative sentences that do not genuinely seek an answer, but instead express evaluation, criticism, or emphasis; the answer is treated as obvious or irrelevant.
    \item \textbf{Leading}: questions that guide the reader or addressee toward a particular answer or interpretation, typically by embedding presuppositions or value judgments. Strongly oriented rhetorical questions are subsumed here.
    \item \textbf{Tag}: declarative statements followed by a short interrogative tag seeking confirmation or alignment.
    \item \textbf{Echo-clarification}: interrogative fragments or sentences that repeat or closely mirror prior wording in order to check understanding, register surprise, or distance oneself from a quoted formulation.
\end{itemize}

\subsection{Automatic detection of interrogative stances}
\label{subsec:automatic_interrogative}

Our goal is to automatically detect interrogative stances and assign stance types to all sentences in more than one million French-language news articles. We adopt a teacher–student pipeline in which a large language model provides pseudo-labels on a subset of sentences, and a French transformer model, CamemBERT-large~\cite{camembert}, is then fine-tuned on these pseudo-labels and applied at scale.

Articles are segmented into sentences, and during model training and inference each sentence is represented together with a short local context window (neighboring sentences from the same article). This preserves local discourse cues that are often crucial for interpreting interrogative force, while still allowing sentence-level labels. Full preprocessing details are given in Appendix~\ref{app:method_details}.

To obtain high-quality pseudo-labels, we apply a French-capable LLM, Qwen3-30B-A3B-Instruct~\cite{qwen3technicalreport}, to a subset of sentences detected by a high-recall heuristic based on question marks, French interrogative patterns, and common question-raising constructions. All candidates, plus a random subsample of non-candidates for calibration, are passed to the LLM together with their local context. The model is prompted to perform a two-stage classification: (i) decide whether the target sentence expresses an interrogative stance (binary), and (ii) for positives, assign one of the six stance types defined above. We retain only high-confidence pseudo-labels for training. Implementation is documented in Appendix~\ref{app:method_details}.

Using these pseudo-labels, we fine-tune two CamemBERT-large classifiers: a \textbf{binary interrogative detector} (interrogative vs.\ non-interrogative) and a \textbf{six-way stance classifier} for interrogative sentences. High-confidence LLM labels are split into training and validation sets with articles kept disjoint across splits, and models are fine-tuned with weighted cross-entropy loss to compensate for label imbalance. Training hyperparameters and architecture details are reported in Appendix~\ref{app:method_details}. On held-out pseudo-labeled validation sets, the binary detector achieves high accuracy (around 0.97), while the stance classifier reaches reasonable six-way performance (macro-F1 around 0.69), providing a usable approximation of pragmatic types for large-scale analysis. When evaluated against our human gold-standard annotations (Section~\ref{subsec:qualitative}), the binary detector reaches an F1 of 0.78 for the interrogative class and the six-way stance classifier macro-F1 of 0.51, which we treat as sufficient for corpus-level analyses rather than sentence-level claims.

At corpus scale, we apply the binary and stance models sequentially to every sentence. We store both predictions and confidence scores for downstream analyses in Section~\ref{sec:results}.

\subsection{Answer identification}
\label{subsec:answer_identification}

To approximate whether interrogative stances are textually resolved within an article, we implement an embedding-based answer search. For each article, we consider as \emph{questions of interest} all sentences predicted as interrogative with sufficient classifier confidence, and group immediately consecutive questions into a single local issue. We then compute sentence embeddings for all sentences in the article using CamemBERT and, for each group of questions, search only among subsequent sentences for the most similar contiguous span of limited length. If the best-scoring span exceeds a similarity threshold, we treat it as the textual answer and record its location and text. Otherwise, the question group is marked as unanswered in that article. This procedure yields an approximate but systematic measure of textual uptake at scale, namely whether an interrogative is followed by a semantically matched answer-like span within the same article. The resulting rates should be read as indicative upper bounds rather than exact estimates of fully resolved answerhood. Exact implementation is detailed in Appendix~\ref{app:method_details}.

\subsection{Entity and topic annotations}
\label{subsec:ner_topic}

To characterize which actors and groups are mentioned in questions and answers, we run named-entity recognition (NER) on the subset of sentences involved in interrogative stances and their detected answer spans. We apply the multilingual GLiNER model~\cite{Gliner} with a coarse, robust label set: \textsc{person}, \textsc{organization}, \textsc{location}, \textsc{nationality or religious or political group}, \textsc{generic social group}, \textsc{public or audience}, and \textsc{event}. For each interrogative sentence, we run GLiNER on a short local context, and for each detected answer span we run it on the answer text. Entities with model scores below 0.5 are discarded. These annotations form the basis for our actor-level analyses of which actors, groups, and places are foregrounded in interrogative and answer contexts. They capture mention patterns rather than direct accountability relations: an entity named near a question is not necessarily the actor being addressed, challenged, or held responsible by that question.

We also derive topics at both article and question level. Using CamemBERT embeddings, we apply BERTopic~\cite{Bertopic} to the combined corpus to obtain unsupervised article topics, and in parallel run BERTopic on interrogative sentences (with their local context) to obtain question-level topics. For interpretability, we manually group the 100 most frequent BERTopic topics into eight broader \emph{meta-topics} by inspecting top keywords and representative articles. Smaller or mixed clusters are assigned to the closest meta-topic on this basis. Algorithmic details are given in Appendix~\ref{app:method_details}.

\subsection{Indices for interrogativity, answerability, dialogicity, and actor profiles}
\label{subsec:indices}

To relate micro-level interrogative practices to broader debates on agenda-setting, gatekeeping, and voice, we derive a set of interpretable indices from the sentence-level predictions, answer spans, and entity annotations. All indices are computed on the QA-enhanced sentence data and then aggregated by article, outlet, country, or topic.

At the article level, we first compute an \emph{interrogative index} that captures how strongly an article structures its discourse through questions. For article $a$, let $S_a$ be the total number of sentences and $Q_a$ the number of sentences classified as interrogative (any of the six types) with sufficient confidence. The interrogative index is
$
ID_a = \frac{Q_a}{S_a},
$
providing a simple measure of how much an article “problematizes” content rather than presenting it purely declaratively.

We then define an \emph{answerability index}:
$
Ans_a = \frac{A_a}{Q_a},
$
where $A_a$ is the number of interrogative sentences in $a$ for which the answer-identification procedure finds a sufficiently similar answer span in the subsequent text. Low $Ans_a$ values indicate orphan questions that are raised but not taken up by the matching procedure in the article, while high values signal a tendency for interrogatives to be followed by semantically matched answer-like spans in-text, linking them more tightly to explanatory or justificatory work.

To distinguish monologic from more dialogic patterns, we compute a \emph{dialogicity} profile. For each answered question, we detect whether the answer span contains markers of direct speech. At the article level, we track the shares of questions that are (i) unanswered; (ii) answered internally (only narrative text); and (iii) answered externally (through direct quotes). The proportion of externally answered questions in an article functions as a proxy for how far answers are delegated to sources rather than supplied solely by the journalist.

Using the NER annotations, we derive actor-level indices. Each interrogative sentence is categorized as \emph{actor-focused} (mentioning one specific person, organization, or location), \emph{group-focused} (mentioning only broad social or public categories such as national/religious/political groups, generic social groups, or audiences), or \emph{issue-focused} (no detected human or collective entities). The relative frequencies of these types provide an \emph{addressivity} profile, indicating whether interrogative contexts are centered on identifiable actors, diffuse publics, or abstract issues.

\subsection{Qualitative analysis}
\label{subsec:qualitative}

We complement the automatic pipeline with a qualitatively annotated subcorpus used for evaluation and for detailed discourse analysis.

\paragraph{Sampling.} We draw a stratified sample of 700 articles from the combined corpus, balancing data sources and interrogative profiles. Using the LLM-based pseudo-labels from the teacher model, we first mark articles as \emph{question-containing} or \emph{no-question}. Among question-containing articles, we compute a dominant pseudo-labeled interrogative stance per article to ensure coverage of the full range of interrogative functions. We then sample equally from the CCNews-derived corpus and the Swiss local corpus, so that half of the manually annotated articles in each subset come from each source. In total, 400 articles (with both question-containing and no-question items) form the main evaluation set for the classification, 100 additional question-containing articles are double-coded by both annotators for inter-annotator agreement, and 200 further question-containing articles (100 per annotator) broaden the qualitative coverage of outlets and interrogative stances.

\paragraph{Coding scheme.} The unit of analysis is a single \emph{interrogative stance}, which may correspond to an explicit interrogative sentence or to an implicit question-introducing construction that pragmatically opens a Question Under Discussion in the sense of Roberts~\cite{roberts2012information}. For each unit, annotators first code the interactional context (interview vs.\ non-interview) and the addressee type (individual, collective actor, audience, self), then the grammatical form (wh-question, polar, alternative, tag, declarative-question, elliptic, indirect), and finally a primary pragmatic function label that maps onto the six stance types used in the automatic classifier (information-seeking, rhetorical, leading, tag, echo-clarification, framing-procedural).

Beyond these micro-pragmatic labels, the scheme captures how questions enact power and voice through four layers: (i) an interpersonal layer (register, inferred status relation, face-threat); (ii) an information layer (inquisitive vs.\ informative vs.\ rhetorical, presuppositions, conduciveness, directive force); (iii) a stance/style layer (evaluative stance, irony markers, performative display); and (iv) an uptake layer (answerhood expectation, resolution status, and whether an answer is realized in the text). On this basis, each interrogative stance is mapped to one or two macro-functional axes: \emph{Authority positioning}, \emph{Framing/agenda-setting}, \emph{Stance/alignment}, \emph{Legitimation}, or \emph{Discursive strategy}. The detailed codebook and decision tree are available at \url{https://gitlab.idiap.ch/socialcomputing/politics-of-questions}. Appendix~\ref{app:eval} reports inter-annotator agreement metrics.

\section{Results}
\label{sec:results}

\subsection{RQ1: Prevalence and distribution of interrogative stances}
\label{subsec:rq1_results}

Across the corpus, interrogative stances are relatively rare but not exceptional. On average, about $2.5\%$ of sentences in an article are classified as interrogative (mean interrogative index $ID_a = 0.025$, SD $0.054$), and the median article contains no interrogative stance at all ($P_{50}=0$, $P_{90}\approx 0.09$). Among interrogative sentences, framing-procedural questions are the dominant type, accounting for just over half of all interrogatives. Taken together, these distributions indicate that in written news interrogatives are used far more often to structure exposition and highlight issues for readers than to pose open information requests to specific actors.
Information-seeking questions account for about one fifth, rhetorical questions for roughly one seventh, and echo-clarification questions for just under one tenth, while explicitly leading and tag questions each make up only about $2\%$ of interrogatives (Table~\ref{tab:stance_global}). Given the moderate reliability of the six-way classifier, we interpret these stance-type shares as coarse corpus-level approximations and avoid strong claims about small differences among rare or pragmatically adjacent categories. This prevalence estimate is robust to reasonable decision thresholds: varying the binary/stance confidence cutoff from 0.6 to 0.8 changes the absolute number of detected interrogatives but keeps mean article-level interrogative density in a narrow band ($ID_a = 0.0236$-$0.0261$, Appendix Table~\ref{tab:confidence_sensitivity}).

\begin{table}
    \centering
    \rowcolors{2}{}{gray!15}
    \begin{tabular}{lrr}
        \toprule
        Stance & $N_Q$ & \% of interrogatives \\
        \midrule
        echo-clarification   &  70{,}281 & 9.2 \\
        framing-procedural   & 396{,}298 & 52.1 \\
        information-seeking  & 153{,}404 & 20.2 \\
        leading              &  16{,}221 & 2.1 \\
        rhetorical           & 107{,}862 & 14.2 \\
        tag                  &  16{,}116 & 2.1 \\
        \bottomrule
    \end{tabular}
    \caption{Global distribution of predicted interrogative stance types in the corpus.}
    \label{tab:stance_global}
\end{table}

\begin{figure*}[t]
    \centering
    \begin{subfigure}{0.49\textwidth}
        \centering
        \includegraphics[width=\linewidth]{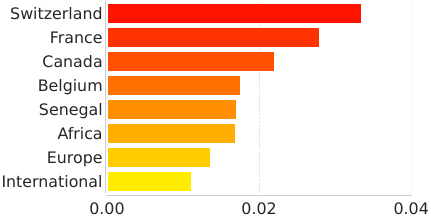}%
        \caption{Mean interrogative index by region.}
        \label{fig:rq1_country_interrog_index}
    \end{subfigure}
    \hfill
    \begin{subfigure}{0.49\textwidth}
        \centering
        \includegraphics[width=\linewidth]{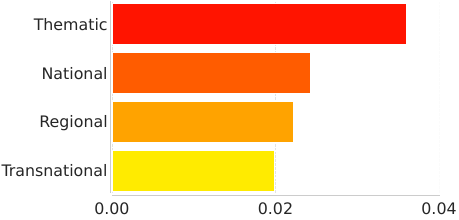}%
        \caption{Mean interrogative index by editorial scale.}
        \label{fig:rq1_scale_interrog_index}
    \end{subfigure}
    \caption{Variation in interrogative density across (a) regions and (b) editorial scales in French-language news (2023–2024).}
    \label{fig:rq1_country_scale_interrog_index}
\end{figure*}

Interrogative density varies systematically across outlets and scales. Grouping by outlet country, Swiss francophone outlets have the highest mean interrogative index ($\bar{I}\approx 0.033$), followed by French outlets ($\bar{I}\approx 0.028$), with Canadian outlets somewhat lower ($\bar{I}\approx 0.022$). Belgian, Senegalese, and pan-African outlets cluster around $0.017$–$0.018$, while outlets coded as broader \enquote{Europe} or \enquote{International} have the lowest interrogative indices ($\bar{I}\approx 0.011$–$0.014$). When grouping by editorial scale, thematic outlets show the highest interrogative index ($\bar{I}\approx 0.036$), national outlets follow ($\bar{I}\approx 0.024$), regional outlets are slightly lower ($\bar{I}\approx 0.022$), and transnational outlets lowest ($\bar{I}\approx 0.020$). Figure~\ref{fig:rq1_country_scale_interrog_index} summarizes these patterns. Overall, interrogative stances are used more intensively in context-rich, domestically anchored coverage and in thematic verticals, and less in transnational or wire-type reporting.

To examine topical variation, we cluster articles with BERTopic and group frequent topics into eight meta-topics (professional sports, national/local politics, geopolitics, local news, lifestyle/entertainment, \emph{faits divers}, business/economy, technology). Three domains stand out as particularly \enquote{question-saturated}: professional sports (mean $ID_a\approx 0.029$ over $\sim 181{,}000$ articles), local news ($\approx 0.030$ over $\sim 39{,}600$ articles), and lifestyle/entertainment ($\approx 0.027$ over $\sim 44{,}500$ articles). Business/economy, geopolitics, and technology are less interrogative (means around $0.018$–$0.022$), with national/local politics and \emph{faits divers} in the middle (around $0.021$–$0.022$, see Appendix Table~\ref{tab:meta_topics}). Stance distributions are broadly similar across topics: framing-procedural questions account for around half to two-thirds of interrogatives everywhere, information-seeking questions are relatively more prominent in \emph{faits divers} and business/economy, while rhetorical and leading questions are somewhat more frequent in national/local politics and geopolitics.

These aggregate patterns match how interrogatives are used in individual articles. In a Canadian explainer on procedures for a one-off family allowance, for instance, the piece is explicitly structured around a sequence of questions (\emph{Who is eligible? How do you apply? What documents are needed?}), each immediately followed by a short answer. Here, interrogatives function as framing-procedural devices that break down a complex policy into manageable steps, illustrating why local and service-oriented topics have relatively high interrogative indices. By contrast, a Swiss commentary on a proposed pension payment is almost entirely declarative, but closes its presentation of the proposal with a single question, \emph{What should we make of this argument~?}, which serves as a rhetorical hinge to orient readers toward the ensuing evaluative discussion. Even in low-interrogative articles, a lone framing question can thus perform macro-structuring work.

\subsection{RQ2: Answerability and dialogicity}
\label{subsec:rq2_results}

\paragraph{Answerability.}

Once interrogative stances appear in news text, they are overwhelmingly taken up in subsequent discourse. Across the corpus, 760{,}182 sentences are classified as interrogative, and for 726{,}911 of them our heuristic identifies a subsequent answer-like span in the same article, corresponding to a 95.6\% rate of heuristic local textual uptake, which we interpret as an upper bound on answerhood rather than as verified resolution in the strict pragmatic sense. At the article level, the median answerability index $Ans_a$ is 1.0 in all outlet groups: in a typical article that contains questions, almost all are followed by some form of semantically matched textual uptake. Answerability varies only modestly across stance types: information-seeking and echo-clarification questions have the highest match rates (97.2\% and 97.4\%), framing-procedural questions the lowest (94.8\%), and rhetorical, leading, and tag questions fall in between (around 95\%, see Appendix Table~\ref{tab:answ_dialog}). Taken together with our qualitative coding, this suggests that even when interrogatives function evaluatively or rhetorically, they are usually embedded in local question-answer sequences rather than left entirely hanging. Because the answer-matching step captures semantic relatedness rather than manually verified resolution, we interpret the 95.6\% figure as a heuristic upper bound on local textual uptake rather than a literal estimate of fully resolved answerhood. Still, this result is not driven by a narrow parameter choice: answerability is unchanged for cosine thresholds from 0.05 to 0.80 (Appendix Table~\ref{tab:answer_sensitivity}), and a manual audit of 50 question groups found clear or partial local answers in 34/40 predicted answered cases, an answer somewhere in the article in 37/40, and confirmed all 10 predicted unanswered cases (Appendix Table~\ref{tab:answer_manual_spotcheck}).

A minority of \enquote{orphan} questions nonetheless remain. For example, in a French local story on new street ashtrays installed to curb cigarette litter, the journalist reports municipal justifications and local associations’ concerns before asking whether more bollard-ashtrays should be added at a specific location. The article ends without answering this question, leaving readers to infer that current measures may be insufficient or inappropriate. Such unresolved, suggestion-like interrogatives correspond closely to the small set of unanswered questions flagged by our automated method.

\paragraph{Dialogicity.}

Although most interrogatives are textually taken up, the form of that uptake differs. Overall, 4.4\% of interrogative sentences remain unanswered, 80.3\% are answered internally by the journalist in narrative text, and 15.4\% are answered through direct quotes or clearly marked reported speech (Appendix Table~\ref{tab:answ_dialog}, bottom panel). In other words, questions in French-language news are overwhelmingly taken up in-text, but in roughly one out of six cases this uptake takes the form of letting a quoted source \enquote{speak back} rather than the journalist alone filling the gap.

\begin{figure}[t]
    \centering
    \includegraphics[width=\linewidth]{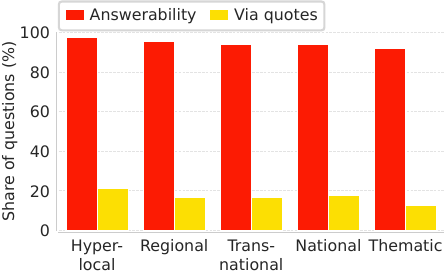}
    \caption{Mean answerability and external dialogicity by editorial scale. Values are shares of all interrogative sentences in each scale.}
    \label{fig:rq2_scale_answerability_dialogicity}
\end{figure}

These shares are relatively stable across countries and outlet scales in terms of answerability, but external dialogicity varies more. Swiss and Belgian outlets show the highest mean fraction of questions answered by quoted speech (around 18\%), followed by French, pan-African, and Senegalese outlets (around 15–16\%), while Canadian and international outlets are somewhat lower (around 13\%). By editorial scale, hyper-local outlets have the highest external dialogicity (20\%), national, regional and transnational outlets follow closely, and thematic outlets (sports, lifestyle, verticals) show the lowest share (around 13\%, Figure~\ref{fig:rq2_scale_answerability_dialogicity}). Together, these results suggest that while questions are almost always resolved, national and regionally rooted outlets more frequently embed answers in dialogic exchanges with sources, whereas thematic verticals rely more heavily on internal narration.

External dialogicity is visible in concrete articles. In a piece on doctored political images and AI-generated \enquote{fake news}, for instance, the journalist raises the issue of where the misleading pictures originated and immediately attributes responsibility in a quoted exchange with a radio host who admits to having created them. Here, an information-seeking question is resolved through an external answer, matching our finding that a substantial share of questions are answered dialogically via quoted sources.

\subsection{RQ3: Who is questioned, and how are interrogatives populated with actors and places?}
\label{subsec:rq3_results}

\paragraph{Where questions are voiced.}

Interrogative stances occur predominantly in journalistic narration rather than as part of direct quotations. Globally, only 7.8\% of interrogative sentences contain quotation markers in the question sentence itself, and 22.3\% if we consider the concatenation of the question and its immediate answer span (figures derived from the same counts as Appendix Table~\ref{tab:answ_dialog}). The remaining three quarters of questions are thus framed in the journalist’s own voice, even when followed by quoted answers. This asymmetry suggests that, within the article text, journalists more often than sources control which questions are explicitly articulated. Sources more often appear as respondents within question-led structures than as initiators of the interrogative agenda. This pattern holds across stance types, but rhetorical, leading, and echo-clarification questions are noticeably more likely to appear inside quoted speech, reflecting their frequent use in talk shows, political interventions, or opinionated statements by sources. Information-seeking and framing-procedural questions, by contrast, are typically voiced by journalists.

\paragraph{Personalization and collectivization.}

Named-entity annotations show that interrogative space appears densely populated with concrete actors and sites. A majority of interrogative sentences mention at least one \textsc{person} (58.5\%), and nearly half mention an \textsc{organization} (43.7\%), almost one third contain a \textsc{location} (31.8\%), and one quarter co-occur with an \textsc{event} entity such as a match, election, or crisis. By contrast, explicit collectivizing references are comparatively rare: only 2.7\% of questions mention a \textsc{public or audience}, 1.3\% a \textsc{nationality or religious or political group}, and 0.3\% a \textsc{generic group}. Interrogative space in this corpus is therefore strongly personalized and institutionalized, with broad social groups and publics appearing much less frequently as explicit addressees. In most cases, questions thus function as organizing devices in discourse \emph{about} actors and places rather than as contributions to an ongoing dialogic exchange with them. Part of this pattern likely reflects standard news sourcing norms, which privilege identifiable elites and institutions over diffuse publics, and the fact that appeals to \enquote{the public} are often left implicit rather than being named as such.

This personalization is high across all countries but varies in degree. Senegalese outlets have the highest personalization index (68.1\% of questions mentioning at least one person), followed by Canadian (60.9\%) and French (59.8\%) outlets. Swiss outlets and Europe-wide or transnational feeds display somewhat lower levels (around 47–50\%), suggesting a slightly less person-centered interrogative style. When grouping by editorial scale, thematic verticals (especially sports and people/lifestyle) show the highest personalization (72.8\% of interrogatives mentioning a person), national outlets are intermediate (55.9\%), and regional and hyper-local outlets are lower (50.1\% and 44.0\%), consistent with a stronger emphasis on institutions, places, and issues than on a small set of named individuals. Collectivization remains a minority phenomenon everywhere (roughly 3–6\% of questions mention a collective), slightly more common in pan-African, Senegalese, and transnational outlets. For brevity, full breakdowns by country and scale are omitted.

\paragraph{Institutional and geographic anchoring by topic.}

Entity annotations also reveal systematic differences in how questions are anchored in institutions, places, and events across topics (Appendix Table~\ref{tab:meta_topics}). In business and economy coverage, interrogatives are especially institution-centered: nearly 58\% of questions mention an organization, compared to 30\% mentioning a location and 25\% an event. Professional sports also shows high organizational anchoring (52\%) and strong event orientation (33\%), with questions frequently tied to specific clubs, competitions, or fixtures. In geopolitics, geographic anchoring is pronounced: almost half of interrogatives mention at least one location and about one third an event, reflecting the centrality of countries, regions, and crises. National/local politics and local news likewise exhibit substantial organizational and locational anchoring, indicating that interrogative contexts in these domains are frequently organized around institutions and localities. By contrast, lifestyle/entertainment and technology topics have lower institutional and geographic anchoring, with interrogatives more often revolving around individuals or generalized situations than around specific organizations or places.

Frequency lists of named entities provide a more concrete picture of who and what dominates interrogative space. Among \textsc{event} labels, major sports competitions and high-salience political or social events are most frequent (\emph{Mercato}, \emph{élections}, \emph{grève}). The \textsc{location} list is headed by national and regional nodes of attention, including \emph{France}, \emph{Paris}, \emph{Ukraine}, \emph{Québec}, \emph{Gaza}, \emph{Suisse}, and \emph{Marseille}. For \textsc{organization}, football clubs and media brands dominate (\emph{Real Madrid}, \emph{Ligue 1}), alongside generic institutional labels (\emph{gouvernement}, \emph{police}, \emph{Union européenne}). Among \textsc{person} entities with proper names, high-profile athletes (Kylian Mbappé, Lionel Messi) and national leaders (Emmanuel Macron, Donald Trump) appear most often, together with a smaller set of recurring political and media figures.
An example illustrates how interrogatives can simultaneously personalize, moralize, and anchor issues geographically. In a French report on protests against Israel’s military actions, a quoted left-wing MP asks, \emph{What is he waiting for before denouncing the Netanyahu government’s war crimes~?}, referring to the French president. The question is embedded in direct speech, targets named elite actors (the president and the Netanyahu government), presupposes culpable delay, and implicitly addresses a wider French public. It exemplifies the pattern whereby rhetorical and leading questions are often voiced by politicians or activists, concentrate on prominent individuals and governments, and combine interrogative form with strong evaluative force.

\subsection{RQ4: Triangulated validation and qualitative insights}

\paragraph{Model evaluation and reliability.}

We assess the reliability of our automatic pipeline by comparing model predictions to the manually annotated gold-standard sample (Section~\ref{subsec:qualitative}). On 8{,}399 gold sentences (both corpora, both annotators), the binary interrogative detector achieves 0.97 accuracy, with precision 0.76, recall 0.80, and F1-score 0.78 for the positive (interrogative) class. On the subset of 516 gold-positive sentences, the six-way stance classifier attains a macro-averaged F1 of 0.51 (micro-F1 0.51). Appendix Table~\ref{tab:model_iaa} summarizes these aggregate figures. These scores indicate that while fine-grained pragmatic distinctions remain challenging, the models reliably capture interrogative vs. non-interrogative status and provide a usable, if imperfect, approximation of stance types for large-scale analysis. Because the student models are trained on high-confidence LLM pseudo-labels, they may also inherit systematic biases from the teacher model, especially for rare or pragmatically adjacent stance types. Appendix Table~\ref{tab:stance_per_class} and the conditional confusion matrix in Figure~\ref{fig:stance_confusion_matrix} further show that the main ambiguities are concentrated among framing-procedural, information-seeking, rhetorical, and leading questions: information-seeking and rhetorical cases are often absorbed into framing-procedural, while leading questions are frequently mapped to rhetorical or framing-procedural. In other words, most residual errors arise among pragmatically adjacent categories rather than between clearly dissimilar question types.

We also measure inter-annotator agreement on a separate set of 98 articles double-coded by both annotators. A fuzzy one-to-one alignment of annotated spans yields an approximate Jaccard overlap of 0.83 over the union of segments. On the 204 matched segments, the two annotators reach 0.84 accuracy on the six-way stance labels, with Cohen’s $\kappa \approx 0.78$. This indicates substantial agreement and confirms that the stance taxonomy is interpretable and can be applied consistently. The gap between human–human and human–model performance is in line with expectations for a nuanced pragmatic task and underscores the value of combining automatic predictions with qualitative analysis. At the same time, the gold annotations should be understood as a carefully constructed reference standard rather than an error-free ground truth: because interrogative stance is context-sensitive and often multifunctional, some model-gold disagreements plausibly reflect borderline cases that admit more than one reasonable reading. We therefore treat stance distributions as approximate and use them primarily for corpus-level patterns, complemented by detailed qualitative analysis. This caution is reinforced by the threshold checks reported above: the main corpus-level patterns are stable even when the absolute number of detected interrogatives varies modestly with stricter confidence cutoffs. This mixed-method, teacher–student design offers a pragmatic compromise: a relatively low-cost way to obtain corpus-scale coverage that remains anchored in a reliable qualitative scheme.

\paragraph{Macro-functions and qualitative insights.}

The qualitative coding shows that interrogatives in French-language news serve a small set of recurrent macro-functions. Across the coded sample, framing and agenda-setting questions dominate (60\% of coded units), followed by authority-positioning interrogatives (30\%). The remaining macro-axes, stance/alignment, discursive strategy, and legitimation each account for only a few percent of questions, but are structurally distinctive. Annotators frequently noted plausible \emph{secondary} axes within a single question, for example when framing questions also expressed stance or legitimation. This multi-functionality suggests that journalistic interrogatives often combine several interactional jobs at once.

Three cross-cutting findings help interpret the large-scale patterns reported above. First, interview contexts act as a powerful normalizing force. Questions addressed to live interlocutors are overwhelmingly information-seeking, non-conducive, and low in face-threat, enacting what Clayman and Heritage term \enquote{soft accountability}: journalists invite explanations or justifications from epistemically privileged sources and almost always receive answers. Outside interviews, by contrast, journalists use interrogatives more freely in a narrative-authority mode, posing questions that they then answer themselves. These questions often include presuppositions and section-opening roles, allowing journalists to display inquiry while retaining interpretive control.

Second, presuppositions are strongly associated with conduciveness, but the qualitative data clarify that it is their \emph{content} that matters. Many framing questions rely on factual presuppositions (e.g. that a statistical decline has occurred) that anchor shared context without steering evaluation, supporting relatively neutral framing. A smaller but important subset combines evaluative presuppositions with mild conduciveness, gently guiding readers toward particular interpretations, especially in hyper-local commentary. This refines classical claims that presuppositions are inherently ideological: in practice they are a flexible resource that can be deployed for both neutral and engaged framing.

Third, hyper-local outlets exhibit an \enquote{engaged but largely neutral} profile. They use more mildly leading or evaluative questions than national outlets, particularly when highlighting local institutional strains or community concerns, yet most of these interrogatives remain low in face-threat and leave space for competing views. This helps explain why, in the aggregate, hyper-local and regional outlets show higher interrogative indices and higher personalization of questions than transnational sources, while still maintaining very high answerability and relatively soft forms of accountability.
Together, these qualitative insights support the validity of our computational indices and show how micro-pragmatic features, presupposition, conduciveness, addressee design, and context, combine to produce the macro-level distributions of interrogativity, answerability, dialogicity, and actor presence observed in French-language news.

\section{Discussion}
\label{sec:discussion}

Our analysis shows that interrogative stances are sparse in French-language news but functionally rich. Although only a small share of sentences are interrogative, they are recruited for a limited and recurrent set of macro-functions: framing and agenda-setting dominate, authority-positioning forms a substantial secondary cluster, and stance/alignment, discursive strategy, and legitimation remain rarer. This pattern is broadly stable across countries, outlet types, and topics, while the frequent presence of secondary macro-axes confirms that individual interrogatives often perform more than one interactional job at once.

In terms of authority and the textual staging of accountability, our findings resonate with and nuance prior work on news interviews and epistemics~\cite{clayman_news_2002,heritage_epistemics_2012}. Authority-positioning interrogatives in our corpus are overwhelmingly \enquote{soft} in Clayman and Heritage’s sense: in interview contexts, journalists typically adopt an epistemically subordinate stance, invite explanations from K$^{+}$ sources, and almost always receive textual uptake, a pattern more consistent with platform-based accountability than with adversarial challenge.
Outside interviews, journalists use interrogatives in a narrative-authority mode, posing questions they then take up themselves and often answer in the surrounding narration, thereby consolidating interpretive control while sustaining a rhetorical appearance of inquiry. High answerability in this sense likely reflects not only accountability dynamics but also the conventions of explanatory and service-oriented digital journalism, where question-led structures are routinely used to organize exposition. Compared to Anglo-American traditions that have documented rising face-threat and negative interrogatives in high-stakes press conferences~\cite{clayman_pressuring_2023}, French-language news in our corpus remains largely consensus-oriented: face-threatening questions are rare, and even hyper-local outlets with somewhat higher conduciveness and leading forms seldom resort to overtly confrontational designs.

Framing interrogatives suggest that questions can serve as a central resource for agenda-setting and issue construction~\cite{entman_framing_1993,van2007constructionist}, but they also complicate simple claims about presuppositions and bias. The large majority of framing questions in our qualitative sample either introduce topics without presuppositions or employ factual presuppositions that anchor shared context without pushing a particular evaluation. A smaller but important subset uses evaluative presuppositions and mild conduciveness to gently steer interpretation, particularly in hyper-local journalism where community-oriented guidance appears more acceptable. Alternative (binary) questions, though rare, illustrate how interrogative form can discretely narrow the interpretive space by restricting readers to two options. These patterns refine Fowler’s~\cite{fowler2013language} view of presuppositions as inherently ideological: in our data, presuppositions are a flexible pragmatic resource that can support neutral or engaged framing depending on what is taken for granted. Our findings on voice and actor salience can likewise be read in relation to classic concerns in media sociology~\cite{shoemaker_gatekeeping_2009,mccombs_evolution_1993}, while adding a pragmatic lens. At scale, interrogatives are heavily populated with named persons, organizations, and places, and a relatively small set of actors and sites, national leaders, key institutions, emblematic locations, recur disproportionately in questions, especially in sports and political coverage. Publics and broad social groups, by contrast, are rarely addressed directly in interrogative form. These mention patterns should be read as indicators of interrogative foregrounding and salience, not as direct measurements of who is actually being held accountable. Qualitatively, we observe two main voice-allocation mechanisms: in interviews, actors are explicitly given the floor to \enquote{speak back} under soft accountability. In non-interview contexts, questions more often organize discourse \emph{about} actors and issues without creating interactional rights to respond. This helps to explain why our dialogicity measures suggest that most questions are followed by textual uptake, often by the journalist, and only a minority of interrogatives are resolved through quoted voices.

Methodologically, the study illustrates how a triangulated, mixed-methods approach can render abstract pragmatic and sociological concepts empirically tractable at scale. Large language model pseudo-labels and fine-tuned CamemBERT classifiers allowed us to detect interrogative stances and approximate their types over 1.2M articles, an embedding-based heuristic provided usable answer spans, and multilingual NER, combined with simple indices, gave a coarse yet informative picture of actor and place salience. The qualitative coding then validated and nuanced these automatic outputs: it confirmed robust patterns (such as interview neutrality constraints and the link between presuppositions and conduciveness) and revealed more subtle distinctions (such as neutral vs.\ evaluative irony, or neutral vs.\ engaged framing) that would be invisible from model outputs alone.

At the same time, our approach has clear limitations. More specifically, the teacher-student setup may propagate some of the pseudo-labeler’s preferences into the final classifiers. Human evaluation mitigates this risk, but does not eliminate it, particularly for infrequent and borderline categories. Our actor-level measures rely on named-entity mentions and quoted-speech markers, so they capture textual foregrounding and voice allocation only imperfectly and should not be read as direct measures of accountability or agenda-setting effects. It is restricted to written French news over a two-year period, stance and NER models are imperfect, especially for rare pragmatic types and marginal actor categories, and the qualitative sample, while theoretically informed, remains relatively small. Future work could extend this framework to broadcast interviews and parliamentary debates, refine actor typing (for instance, distinguishing state, corporate, NGO, and citizen voices more systematically), and replicate the analysis in other languages and media systems. More broadly, the combination of large-scale NLP and detailed pragmatic coding offers a promising route to studying how questions, answers, and the distribution of voice structure contemporary public discourse.

\section{Conclusion}


This paper examined the \emph{politics of questions} in contemporary French-language news through a mixed-methods analysis of more than 1.2 million articles (2023-2024), combining LLM-based pseudo-labeling, fine-tuned neural classifiers, heuristic answer detection, multilingual NER, and a qualitatively coded subcorpus grounded in pragmatic theory.

For \textbf{RQ1}, we find that interrogative stances are quantitatively sparse but highly structured: only a small share of sentences per article are interrogative, yet framing-procedural questions dominate wherever they appear, with information-seeking and echo-clarification questions forming most of the remainder and explicitly leading and tag questions rare. Interrogative density varies systematically across media systems, with Swiss and French outlets and thematic verticals, especially sports, local news, and lifestyle, relying more on question-based structuring than business, geopolitics, or technology. For \textbf{RQ2}, our heuristic identifies an answer-like textual continuation for the vast majority of questions, but this uptake is mostly monologic: questions are usually followed by the journalist’s own narrative continuation, and only a minority are resolved through quoted speech. These rates should be read as upper-bound indicators of textual uptake rather than fully verified answerhood. National and some regional outlets are somewhat more dialogic than thematic or transnational sources, but the overall pattern of high uptake and limited external dialogicity is robust.

Turning to \textbf{RQ3}, our NER-based analysis reveals that interrogative space is densely populated with named persons, organizations, locations, and emblematic events, while explicit publics and generic social groups are rarely addressed. Personalization is especially strong in Senegalese and Canadian outlets and in thematic verticals, and somewhat lower in Swiss and hyper-local outlets, which more often foreground institutions and places. Across topics, interrogatives in business and sports are strongly anchored in organizations and events, while geopolitical and local-political questions foreground locations. Elites and key institutions thus occupy the center of both questioning and response. Finally, for \textbf{RQ4}, our teacher–student pipeline yields high reliability for the binary distinction between interrogative and non-interrogative sentences and usable, though imperfect, performance for six-way stance types. Qualitative coding organized around macro-functional axes is essential for interpreting these outputs, distinguishing neutral from engaged framing and clarifying how presuppositions, conduciveness, and face-threat shape the interactional work of questions.

These findings speak to debates in journalism studies, pragmatics, and computational social science. In our corpus, interrogative stances appear intermittently and are most often mobilized to structure explanation or mark issues for consideration rather than to stage overt confrontation. Authority-positioning interrogatives tend to take \enquote{soft} forms: in interviews, journalists invite explanations from epistemically privileged sources, while in non-interview contexts they frequently pose questions they themselves go on to answer, combining displays of inquiry with a strong degree of narrative control. Framing questions, especially those with evaluative presuppositions or binary alternatives, show how interrogatives can subtly steer interpretation, yet many presuppositions function in a relatively neutral way, anchoring shared context without straightforwardly imposing ideological commitments. From a gatekeeping perspective, the strong personalization and institutionalization of interrogative discourse, combined with relatively modest direct address to publics, are consistent with the idea that questions tend to foreground actors already central to the news agenda and to allocate voice unevenly across article texts. Methodologically, the study demonstrates that integrating pragmatic concepts such as interrogative stance, presupposition, answerhood, and dialogicity into large-scale NLP can enrich computational accounts of agenda-setting and voice.

The study also has limitations. It is restricted to written French-language news over a two-year period, and several automatic components remain approximate: stance classification is imperfect, the NER schema is coarse, and the analysis relies on textual traces alone rather than the prosodic and interactional cues available in broadcast or face-to-face settings. Our indices of interrogativity, answerability, dialogicity, and voice should therefore be read as interpretable proxies rather than precise reconstructions of interactional practice. Future work could extend this framework to other media genres, languages, and historical periods while refining actor typing and dialogic attribution.

\section*{Acknowledgments}
This work was supported by the EU Horizon Europe program through the ELIAS project (No. 101120237) and by the EU Horizon Europe program, Marie Skłodowska-Curie Actions (MSCA), alignAI project (No.  101169473).
We thank Isabelle Segarini and Frédéric Keller (ESH Médias - \url{https://www.eshmedias.ch/}) for discussions and support with data news provision.

\begingroup
\setlength{\bibsep}{0.25pt plus 0.5ex}
\small
\bibliography{aaai2026}

@book{hindman_internet_2018,
	title = {The {Internet} {Trap}: {How} the {Digital} {Economy} {Builds} {Monopolies} and {Undermines} {Democracy}},
	isbn = {978-0-691-15926-3},
	shorttitle = {The {Internet} {Trap}},
	url = {https://www.jstor.org/stable/j.ctv36zrf8},
	publisher = {Princeton University Press},
	author = {Hindman, Matthew},
	year = {2018},
	doi = {10.2307/j.ctv36zrf8},
}

@misc{FOEG_Yearbook_2024,
  author = {Udris, Linards and F{\"u}rst, Silke and Eisenegger, Mark},
  title = {Are news from public service media crowding out private news media? Usage and willingness to pay in Switzerland},
  year = {2024}
}

@article{abernathy_news_2023,
	title = {News {Deserts}: {A} {Research} {Agenda} for {Addressing} {Disparities} in the {United} {States}},
	volume = {11},
	issn = {2183-2439},
	shorttitle = {News {Deserts}},
	url = {https://www.cogitatiopress.com/mediaandcommunication/article/view/6728},
	doi = {10.17645/mac.v11i3.6728},
	language = {en},
	number = {3},
	journal = {Media and Communication},
	author = {Abernathy, Penelope Muse},
	month = sep,
	year = {2023},
	pages = {290--292},
}

@article{vogler_investigating_2023,
  title = {Investigating News Deserts on the Content Level: Geographical Diversity in Swiss News Media},
  journal = {Media and Communication},
  author = {Vogler, D. and Weston, M. and Udris, L.},
  year = {2023}
}

@article{shaker_dead_2014,
	title = {Dead {Newspapers} and {Citizens}’ {Civic} {Engagement}},
	volume = {31},
	issn = {1058-4609},
	url = {https://doi.org/10.1080/10584609.2012.762817},
	doi = {10.1080/10584609.2012.762817},
	number = {1},
	journal = {Political Communication},
	author = {Shaker, Lee},
	month = jan,
	year = {2014},
	note = {Publisher: Routledge
\_eprint: https://doi.org/10.1080/10584609.2012.762817},
	pages = {131--148},
}

@article{barclay_local_2025,
	title = {Local news as political institution and the repercussions of ‘news deserts’: {A} qualitative study of seven {UK} local areas},
	volume = {26},
	issn = {1464-8849},
	shorttitle = {Local news as political institution and the repercussions of ‘news deserts’},
	url = {https://doi.org/10.1177/14648849241272255},
	doi = {10.1177/14648849241272255},
	language = {EN},
	number = {9},
	journal = {Journalism},
	author = {Barclay, Steven and Barnett, Steven and Moore, Martin and Townend, Judith},
	month = sep,
	year = {2025},
	note = {Publisher: SAGE Publications},
	pages = {1803--1821},
}

@book{pickard_democracy_2020,
	title = {Democracy without {Journalism}?: {Confronting} the {Misinformation} {Society}},
	shorttitle = {Democracy without {Journalism}?},
	author = {Pickard, Victor},
	year = {2020},
	doi = {10.1093/oso/9780190946753.001.0001},
}

@book{clayman_news_2002,
	series = {Studies in {Interactional} {Sociolinguistics}},
	title = {The {News} {Interview}: {Journalists} and {Public} {Figures} on the {Air}},
	shorttitle = {The {News} {Interview}},
	publisher = {Cambridge University Press},
	author = {Clayman, Steven and Heritage, John},
	year = {2002},
	doi = {10.1017/CBO9780511613623},
}

@incollection{heritage_epistemics_2012,
	title = {Epistemics in {Conversation}},
	url = {https://onlinelibrary.wiley.com/doi/abs/10.1002/9781118325001.ch18},
	language = {en},
	booktitle = {The {Handbook} of {Conversation} {Analysis}},
	publisher = {John Wiley \& Sons, Ltd},
	author = {Heritage, John},
	year = {2012},
	doi = {10.1002/9781118325001.ch18},
	pages = {370--394},
}

@book{ilie_what_1994,
	title = {What {Else} {Can} {I} {Tell} {You}?: {A} {Pragmatic} {Study} of {English} {Rhetorical} {Questions} as {Discursive} and {Argumentative} {Acts}},
	shorttitle = {What {Else} {Can} {I} {Tell} {You}?},
	language = {en},
	publisher = {Almqvist \& Wiksell International},
	author = {Ilie, Cornelia},
	year = {1994},
}

@article{frank_you_1990,
	title = {You call that a rhetorical question?: {Forms} and functions of rhetorical questions in conversation},
	volume = {14},
	issn = {0378-2166},
	shorttitle = {You call that a rhetorical question?},
	url = {https://www.sciencedirect.com/science/article/pii/037821669090003V},
	doi = {10.1016/0378-2166(90)90003-V},
	number = {5},
	journal = {Journal of Pragmatics},
	author = {Frank, Jane},
	month = oct,
	year = {1990},
	pages = {723--738},
}

@book{ginzburg_interactive_2015,
	address = {Oxford, GB},
	title = {The {Interactive} {Stance}: {Meaning} for {Conversation}},
	shorttitle = {The {Interactive} {Stance}},
	publisher = {Oxford University Press},
	author = {Ginzburg, Jonathan},
	year = {2015},
}

@article{van_rooy_questioning_2003,
	title = {Questioning to {Resolve} {Decision} {Problems}},
	volume = {26},
	issn = {1573-0549},
	url = {https://doi.org/10.1023/B:LING.0000004548.98658.8f},
	doi = {10.1023/B:LING.0000004548.98658.8f},
	language = {en},
	number = {6},
	journal = {Linguistics and Philosophy},
	author = {van Rooy, Robert},
	month = dec,
	year = {2003},
	pages = {727--763},
}

@article{greatbatch_turn-taking_1988,
	title = {A {Turn}-{Taking} {System} for {British} {News} {Interviews}},
	issn = {0047-4045},
	url = {https://www.jstor.org/stable/4167953},
	journal = {Language in Society},
	author = {Greatbatch, David},
	year = {1988},
}

@book{hallin_comparing_2004,
  title = {Comparing {Media} {Systems}: {Three} {Models} of {Media} and {Politics}},
  author = {Hallin, Daniel C. and Mancini, Paolo},
  year = {2004}
}

@incollection{straubhaar_cultural_2021,
	title = {Cultural {Proximity}},
	booktitle = {The {Routledge} {Handbook} of {Digital} {Media} and {Globalization}},
	publisher = {Routledge},
	author = {Straubhaar, Joseph},
	year = {2021},
}

@article{mccombs_evolution_1993,
	title = {The {Evolution} of {Agenda}-{Setting} {Research}: {Twenty}-{Five} {Years} in the {Marketplace} of {Ideas}},
	volume = {43},
	issn = {1460-2466},
	shorttitle = {The {Evolution} of {Agenda}-{Setting} {Research}},
	doi = {10.1111/j.1460-2466.1993.tb01262.x},
	language = {en},
	number = {2},
	journal = {Journal of Communication},
	author = {McCombs, Maxwell E. and Shaw, Donald L.},
	year = {1993},
	pages = {58--67},
}

@article{hudson_meaning_1975,
	title = {The {Meaning} of {Questions}},
	doi = {10.2307/413148},
	language = {en},
	journal = {Language},
	author = {Hudson, Richard A.},
	month = mar,
	year = {1975},
}

@book{stokhof_jeroen_nodate,
    author = {Groenendijk, Jeroen and Stokhof, Martin},
    year = {1984},
    month = {01},
    pages = {},
    title = {Studies on the Semantics of Questions and the Pragmatics of Answers},
    journal = {Varieties of Formal Semantics}
}

@book{ciardelli_questions_2016,
	address = {S.l},
	title = {Questions in logic},
	isbn = {978-90-6464-977-6},
	language = {en},
	publisher = {s.n.},
	author = {Ciardelli, Ivano A.},
	year = {2016},
	note = {OCLC: 945552181},
}

@book{ciardelli_inquisitive_2018,
	title = {Inquisitive {Semantics}},
	isbn = {978-0-19-881478-8},
	url = {https://library.oapen.org/handle/20.500.12657/25127},
	language = {English},
	publisher = {Oxford University Press},
	author = {Ciardelli, Ivano and Groenendijk, Jeroen and Roelofsen, Floris},
	year = {2018},
	doi = {10.1093/oso/9780198814788.001.0001},
}

@article{farkas_reacting_2010,
	title = {On {Reacting} to {Assertions} and {Polar} {Questions}},
	volume = {27},
	issn = {0167-5133},
	url = {https://doi.org/10.1093/jos/ffp010},
	doi = {10.1093/jos/ffp010},
	number = {1},
	journal = {Journal of Semantics},
	author = {Farkas, Donka F. and Bruce, Kim B.},
	month = feb,
	year = {2010},
	pages = {81--118},
}

@article{braun_implicating_2011,
	title = {Implicating {Questions}},
	copyright = {http://doi.wiley.com/10.1002/tdm\_license\_1.1},
	issn = {02681064},
	url = {https://onlinelibrary.wiley.com/doi/10.1111/j.1468-0017.2011.01431.x},
	doi = {10.1111/j.1468-0017.2011.01431.x},
	language = {en},
	journal = {Mind \& Language},
	author = {Braun, David},
	month = nov,
	year = {2011},
}

@article{stivers_mobilizing_2010,
	title = {Mobilizing {Response}},
	volume = {43},
	issn = {0835-1813},
	doi = {10.1080/08351810903471258},
	number = {1},
	journal = {Research on Language and Social Interaction},
	author = {Stivers, Tanya and Rossano, Federico},
	month = feb,
	year = {2010},
}

@article{clayman_pressuring_2023,
	title = {Pressuring the {President}: {Changing} language practices and the growth of political accountability},
	volume = {207},
	issn = {0378-2166},
	shorttitle = {Pressuring the {President}},
	url = {https://www.sciencedirect.com/science/article/pii/S0378216623000206},
	doi = {10.1016/j.pragma.2023.01.014},
	journal = {Journal of Pragmatics},
	author = {Clayman, Steven E. and Heritage, John},
	month = apr,
	year = {2023},
	pages = {62--74},
}

@book{gans_deciding_2004,
	title = {Deciding {What}'s {News}: {A} {Study} of {CBS} {Evening} {News}, {NBC} {Nightly} {News}, {Newsweek}, and {Time}},
	isbn = {978-0-8101-2237-6},
	publisher = {Northwestern University Press},
	author = {Gans, Herbert J.},
	year = {2004},
}

@book{shoemaker_gatekeeping_2009,
	address = {New York},
	title = {Gatekeeping {Theory}},
	isbn = {978-0-203-93165-3},
	publisher = {Routledge},
	author = {Shoemaker, Pamela J. and Vos, Timothy},
	month = sep,
	year = {2009},
	doi = {10.4324/9780203931653},
}

@article{entman_framing_1993,
	title = {Framing: {Toward} {Clarification} of a {Fractured} {Paradigm}},
	volume = {43},
	copyright = {http://doi.wiley.com/10.1002/tdm\_license\_1.1},
	issn = {0021-9916, 1460-2466},
	shorttitle = {Framing},
	url = {https://academic.oup.com/joc/article/43/4/51-58/4160153},
	doi = {10.1111/j.1460-2466.1993.tb01304.x},
	language = {en},
	number = {4},
	journal = {Journal of Communication},
	author = {Entman, Robert M.},
	month = dec,
	year = {1993},
	pages = {51--58},
}

@article{ranganath_identifying_2016,
	title = {Identifying {Rhetorical} {Questions} in {Social} {Media}},
	volume = {10},
	doi = {10.1609/icwsm.v10i1.14771},
	language = {en},
	journal = {Proceedings of the International AAAI Conference on Web and Social Media},
	author = {Ranganath, Suhas and Hu, Xia and Tang, Jiliang and Wang, Suhang and Liu, Huan},
	year = {2016},
}

@inproceedings{oraby_are_2017,
	title = {Are you serious?: {Rhetorical} {Questions} and {Sarcasm} in {Social} {Media} {Dialog}},
	shorttitle = {Are you serious?},
	doi = {10.18653/v1/W17-5537},
	booktitle = {Proceedings of the 18th {SIGdial} {Meeting} on {Discourse} and {Dialogue}},
	publisher = {Association for Computational Linguistics},
	author = {Oraby, Shereen and Harrison, Vrindavan and Misra, Amita and Riloff, Ellen and Walker, Marilyn},
	year = {2017},
}

@inproceedings{kwak_systematic_2020,
	title = {A {Systematic} {Media} {Frame} {Analysis} of 1.5 {Million} {New} {York} {Times} {Articles} from 2000 to 2017},
	doi = {10.1145/3394231.3397921},
	booktitle = {Proceedings of the 12th {ACM} {Conference} on {Web} {Science}},
	publisher = {Association for Computing Machinery},
	author = {Kwak, Haewoon and An, Jisun and Ahn, Yong-Yeol},
	month = jul,
	year = {2020},
}

@article{dehler-holland_topic_2021,
	title = {Topic {Modeling} {Uncovers} {Shifts} in {Media} {Framing} of the {German} {Renewable} {Energy} {Act}},
	volume = {2},
	issn = {2666-3899},
	url = {https://www.sciencedirect.com/science/article/pii/S2666389920302336},
	doi = {10.1016/j.patter.2020.100169},
	number = {1},
	journal = {Patterns},
	author = {Dehler-Holland, Joris and Schumacher, Kira and Fichtner, Wolf},
	month = jan,
	year = {2021},
	pages = {100169},
}

@article{guo_framing_2011,
	title = {Framing distance: local vs. non-local news in {Hong} {Kong} press},
	doi = {10.1080/17544750.2011.544080},
	journal = {Chinese Journal of Communication},
	author = {Guo, Steve},
	year = {2011},
}

@article{shi_examining_2023,
	title = {Examining the {Intermedia} {Agenda} {Setting} {Effects} amid the {Changsheng} {Vaccine} {Crisis}: {A} {Computational} {Approach}},
	shorttitle = {Examining the {Intermedia} {Agenda} {Setting} {Effects} amid the {Changsheng} {Vaccine} {Crisis}},
	doi = {10.3390/ijerph20054052},
	journal = {International Journal of Environmental Research and Public Health},
	author = {Shi, Jian and Wang, Hanxiao},
	month = feb,
	year = {2023},
	pages = {4052},
}

@article{nguyen_exploring_2024,
	title = {Exploring the {Value} of {Computational} {Methods} for {Metajournalistic} {Discourse}: {The} {Example} of {COVID}-19 {Reporting} in {Dutch} {Newspapers}},
	volume = {25},
	issn = {1461-670X},
	shorttitle = {Exploring the {Value} of {Computational} {Methods} for {Metajournalistic} {Discourse}},
	url = {https://doi.org/10.1080/1461670X.2024.2358118},
	doi = {10.1080/1461670X.2024.2358118},
	number = {10},
	journal = {Journalism Studies},
	author = {Nguyen, Dennis and van Es, Karin},
	month = jul,
	year = {2024},
	note = {Publisher: Routledge
\_eprint: https://doi.org/10.1080/1461670X.2024.2358118},
	pages = {1160--1181},
}

@article{olsen_triangulation_nodate,
	title = {Triangulation in {Social} {Research}:  {Qualitative} and {Quantitative} {Methods} {Can} {Really} {Be} {Mixed}},
	language = {en},
	author = {Olsen, Wendy},
    year = {2004}
}

@article{nelson_computational_2020,
	title = {Computational {Grounded} {Theory}: {A} {Methodological} {Framework}},
	volume = {49},
	issn = {0049-1241},
	shorttitle = {Computational {Grounded} {Theory}},
	url = {https://doi.org/10.1177/0049124117729703},
	doi = {10.1177/0049124117729703},
	language = {EN},
	number = {1},
	journal = {Sociological Methods \& Research},
	author = {Nelson, Laura K.},
	month = feb,
	year = {2020},
	note = {Publisher: SAGE Publications Inc},
	pages = {3--42},
}

@article{gilardi_chatgpt_2023,
	title = {{ChatGPT} outperforms crowd workers for text-annotation tasks},
	volume = {120},
	url = {https://www.pnas.org/doi/10.1073/pnas.2305016120},
	doi = {10.1073/pnas.2305016120},
	number = {30},
	journal = {Proceedings of the National Academy of Sciences},
	author = {Gilardi, Fabrizio and Alizadeh, Meysam and Kubli, Maël},
	month = jul,
	year = {2023},
	note = {Publisher: Proceedings of the National Academy of Sciences},
	pages = {e2305016120},
}

@article{roberts2012information,
  title={Information structure: Towards an integrated formal theory of pragmatics},
  author={Roberts, Craige},
  journal={Semantics and pragmatics},
  volume={5},
  pages={6--1},
  year={2012}
}

@article{van2007constructionist,
  title={The constructionist approach to framing: Bringing culture back in},
  author={Van Gorp, Baldwin},
  journal={Journal of communication},
  year={2007},
  publisher={Oxford University Press}
}

@book{fowler2013language,
  title={Language in the News: Discourse and Ideology in the Press},
  author={Fowler, Roger},
  year={2013},
  publisher={Routledge}
}

@misc{CCNewsHamborg,
  author = {CommonCrawl},
  title = {Common Crawl Corpus},
  year = {2024}
}

@article{Bros_Gatica-Perez_2025, title={The Suisse Romande Local News Dataset}, volume={19}, url={https://ojs.aaai.org/index.php/ICWSM/article/view/35942}, DOI={10.1609/icwsm.v19i1.35942},  number={1}, journal={Proceedings of the International AAAI Conference on Web and Social Media}, author={Bros, Victor and Gatica-Perez, Daniel}, year={2025}, month={Jun.}, pages={2396-2401} }

@misc{Gliner,
  title = {{GLiNER}: Generalist Model for Named Entity Recognition using Bidirectional Transformer},
  author = {Zaratiana, U. and Tomeh, N. and Holat, P. and Charnois, T.},
  year = {2023}
}

@misc{camembert,
      title={CamemBERT 2.0: A Smarter French Language Model Aged to Perfection}, 
      author={Wissam Antoun and Francis Kulumba and Rian Touchent and Éric de la Clergerie and Benoît Sagot and Djamé Seddah},
      year={2024},
}

@misc{Bertopic,
  title = {{BERTopic}: Neural topic modeling with a class-based {TF}-{IDF} procedure},
  author = {Grootendorst, M.},
  year = {2022}
}

@inproceedings{qi-etal-2020-stanza,
    title = "{S}tanza: A Python Natural Language Processing Toolkit for Many Human Languages",
    author = "Qi, Peng  and
      Zhang, Yuhao  and
      Zhang, Yuhui  and
      Bolton, Jason  and
      Manning, Christopher D.",
    booktitle = "Proceedings of the 58th Annual Meeting of the Association for Computational Linguistics: System Demonstrations",
    year = "2020",
    publisher = "Association for Computational Linguistics",
    doi = "10.18653/v1/2020.acl-demos.14",
}

@misc{qwen3technicalreport,
      title={Qwen3 Technical Report}, 
      author={Qwen Team},
      year={2025},
      eprint={2505.09388},
      archivePrefix={arXiv},
      primaryClass={cs.CL},
      url={https://arxiv.org/abs/2505.09388}, 
}

@article{athanasiadou_modes_1991,
author = {Athanasiadou, Angeliki},
year = {1991},
month = {03},
pages = {},
title = {The Discourse Function Of Questions},
volume = {1},
journal = {Pragmatics; Vol 1, No 1 (1991)},
doi = {10.1075/prag.1.1.02ath}
}
\endgroup

\clearpage

\setcounter{secnumdepth}{1}
\renewcommand\thesection{\Alph{section}}

\section*{Paper Checklist}

\begin{enumerate}

\item For most authors...
\begin{enumerate}
    \item  Would answering this research question advance science without violating social contracts, such as violating privacy norms, perpetuating unfair profiling, exacerbating the socio-economic divide, or implying disrespect to societies or cultures?
    \answerYes{Yes, and the work analyzes publicly available news articles and reports only aggregate statistics about outlets, topics, and actors, with no collection of private or user-generated data.}
  \item Do your main claims in the abstract and introduction accurately reflect the paper's contributions and scope?
    \answerYes{Yes, and the Abstract and Introduction state the mixed-methods design, the corpus scope, and the contributions without overclaiming.}
   \item Do you clarify how the proposed methodological approach is appropriate for the claims made? 
    \answerYes{Yes, and we explain why combining large-scale computational analysis with qualitative coding is suitable for our research questions in the Methodology and Results sections.}
   \item Do you clarify what are possible artifacts in the data used, given population-specific distributions?
    \answerYes{Yes, and we describe corpus composition (countries, outlet types, editorial scales, time window) and note coverage and representativeness limitations, for example the focus on French-language digital news and on specific outlet sets.}
  \item Did you describe the limitations of your work?
    \answerYes{Yes, and we discuss limitations in the Discussion and Conclusion.}
  \item Did you discuss any potential negative societal impacts of your work?
    \answerNo{No, because the paper presents very limited risks in this regard. The contributions and the results are presented in aggregations.}
      \item Did you discuss any potential misuse of your work?
    \answerNo{No, because the paper presents very limited risks in this regard.}
    \item Did you describe steps taken to prevent or mitigate potential negative outcomes of the research, such as data and model documentation, data anonymization, responsible release, access control, and the reproducibility of findings?
    \answerYes{Yes, and the code for reproducibility are available at \url{https://gitlab.idiap.ch/socialcomputing/politics-of-questions}.}
  \item Have you read the ethics review guidelines and ensured that your paper conforms to them?
    \answerYes{Yes, and the study has been designed to conform to ethics guidance as we understand it.}
\end{enumerate}

\item Additionally, if your study involves hypotheses testing...
\begin{enumerate}
  \item Did you clearly state the assumptions underlying all theoretical results?
    \answerNA{NA}
  \item Have you provided justifications for all theoretical results?
    \answerNA{NA}
  \item Did you discuss competing hypotheses or theories that might challenge or complement your theoretical results?
    \answerNA{NA}
  \item Have you considered alternative mechanisms or explanations that might account for the same outcomes observed in your study?
    \answerNA{NA}
  \item Did you address potential biases or limitations in your theoretical framework?
    \answerNA{NA}
  \item Have you related your theoretical results to the existing literature in social science?
    \answerNA{NA}
  \item Did you discuss the implications of your theoretical results for policy, practice, or further research in the social science domain?
    \answerNA{NA}
\end{enumerate}

\item Additionally, if you are including theoretical proofs...
\begin{enumerate}
  \item Did you state the full set of assumptions of all theoretical results?
    \answerNA{NA}
	\item Did you include complete proofs of all theoretical results?
    \answerNA{NA}
\end{enumerate}

\item Additionally, if you ran machine learning experiments...
\begin{enumerate}
  \item Did you include the code, data, and instructions needed to reproduce the main experimental results (either in the supplemental material or as a URL)?
    \answerYes{Due to news copyright constraints, we do not redistribute full-text articles. Code and derived non-copyright-restricted artifacts are available at \url{https://gitlab.idiap.ch/socialcomputing/politics-of-questions}}
  \item Did you specify all the training details (e.g., data splits, hyperparameters, how they were chosen)?
    \answerYes{Yes, and we describe model architectures, training/validation splits, and hyperparameters in the Methodology section.}
     \item Did you report error bars (e.g., with respect to the random seed after running experiments multiple times)?
    \answerNo{No, because we report single-run metrics for each configuration and did not repeat training with multiple random seeds.}
	\item Did you include the total amount of compute and the type of resources used (e.g., type of GPUs, internal cluster, or cloud provider)?
    \answerNo{No, because the current draft does not yet specify hardware details or total compute. Time constraints for the publication prevented a reliable estimation of the compute resources.}
     \item Do you justify how the proposed evaluation is sufficient and appropriate to the claims made? 
    \answerYes{Yes, and we justify our evaluation via a manually annotated gold-standard subset, inter-annotator agreement analysis, and model-human comparisons, and we argue that the resulting performance is adequate for aggregate, corpus-level analyses.}
     \item Do you discuss what is \enquote{the cost} of misclassification and fault (in)tolerance?
    \answerYes{Yes, and we explain that stance and NER errors limit fine-grained interpretation, so our indices should be treated as approximate proxies rather than precise reconstructions, and we discuss these limitations in the Methodology, Results, and Discussion.}
  
\end{enumerate}

\item Additionally, if you are using existing assets (e.g., code, data, models) or curating/releasing new assets, \textbf{without compromising anonymity}...
\begin{enumerate}
  \item If your work uses existing assets, did you cite the creators?
    \answerYes{Yes, and we cite the creators of the news corpora and the main NLP models and toolkits we use (CCNews, the Suisse Romande corpus, CamemBERT, BERTopic, GLiNER, stanza, and the LLM teacher model).}
  \item Did you mention the license of the assets?
    \answerNo{No, because we currently describe data sources and tools as they recommend to be cited but do not yet explicitly state licenses.}
  \item Did you include any new assets in the supplemental material or as a URL?
    \answerYes{We release code and derived artifacts at \url{https://gitlab.idiap.ch/socialcomputing/politics-of-questions}, but not the reconstructed full-text corpus due to copyright restrictions.}
  \item Did you discuss whether and how consent was obtained from people whose data you're using/curating?
    \answerNo{No, because we exclusively use publicly available news articles.}
  \item Did you discuss whether the data you are using/curating contains personally identifiable information or offensive content?
    \answerYes{Yes, and we have not encountered such content during the development of the project. Though the corpus may contain personal names and potentially sensitive content as typical in news reporting. We analyze and report only in aggregate, do not redistribute full text, and avoid exposing individuals beyond what is already public.}
\item If you are curating or releasing new datasets, did you discuss how you intend to make your datasets FAIR (see FORCE11 (2020))?
\answerNA{NA}
\item If you are curating or releasing new datasets, did you create a Datasheet for the Dataset (see Gebru et al. (2021))? 
\answerNA{NA}
\end{enumerate}

\item Additionally, if you used crowdsourcing or conducted research with human subjects, \textbf{without compromising anonymity}...
\begin{enumerate}
  \item Did you include the full text of instructions given to participants and screenshots?
    \answerNA{NA}
  \item Did you describe any potential participant risks, with mentions of Institutional Review Board (IRB) approvals?
    \answerNA{NA}
  \item Did you include the estimated hourly wage paid to participants and the total amount spent on participant compensation?
    \answerNA{NA}
   \item Did you discuss how data is stored, shared, and deidentified?
   \answerNA{NA}
\end{enumerate}

\end{enumerate}

\clearpage
\appendix

\setcounter{secnumdepth}{1}
\renewcommand\thesection{\Alph{section}}

\section{Outlets with their ontologies}
\label{app:data}

Table~\ref{tab:outlets_ontology} lists all outlets included in the corpus, with article counts for 2023–2024 and the manually coded outlet ontologies used in the analyses (primary country/region and editorial scale/type).

\begin{table*}[t]
    \centering
    \rowcolors{2}{}{gray!15}
    \begin{tabular}{l r l l l}
        \toprule
        Source & Articles & Country/region & Scale & Type \\
        \midrule
        arcinfo.ch                    & 11{,}204  & Switzerland          & Hyper-local   & general      \\
        footmercato.net       & 35{,}579  & France               & Thematic      & sports       \\
        fr.allafrica.com      &116{,}680  & Africa (pan-African) & Transnational & general      \\
        francetvinfo.fr       & 39{,}709  & France               & National      & general      \\
        gala.fr               & 32{,}556  & France               & Thematic      & celebrity    \\
        ici.radio-canada.ca   & 50{,}976  & Canada               & National      & general      \\
        lacote.ch                    &  6{,}145  & Switzerland          & Hyper-local   & general      \\
        la-croix.com          & 19{,}486  & France               & National      & general      \\
        ladepeche.fr          &176{,}974  & France               & Regional      & general      \\
        lapresse.ca           & 54{,}136  & Canada               & National      & general      \\
        le10sport.com         & 61{,}404  & France               & Thematic      & sports       \\
        lefigaro.fr           & 43{,}444  & France               & National      & general      \\
        lelezard.com          & 23{,}594  & International        & Transnational & wire         \\
        lenouvelliste.ch                    &  9{,}774  & Switzerland          & Hyper-local   & general      \\
        lequipe.fr            & 51{,}389  & France               & Thematic      & sports       \\
        lindependant.fr       & 60{,}960  & France               & Regional      & general      \\
        midilibre.fr          & 89{,}704  & France               & Regional      & general      \\
        news-24.fr            &123{,}725  & France               & Transnational & wire         \\
        rtl.be                & 82{,}375  & Belgium              & National      & general      \\
        seneweb.com           & 27{,}892  & Senegal              & National      & general      \\
        tf1info.fr            & 42{,}118  & France               & National      & general      \\
        voici.fr              & 32{,}335  & France               & Thematic      & celebrity    \\
        watson.ch             & 16{,}127  & Switzerland          & National      & general      \\
        zonebourse.com        & 30{,}120  & Europe               & Transnational & business      \\
        \bottomrule
    \end{tabular}
    \caption{Outlets in the corpus with total article counts (2023-mid-2024) and outlet ontologies.}
    \label{tab:outlets_ontology}
\end{table*}

\section{Additional implementation details}
\label{app:method_details}

For all the stochastic runs described below, we used a fixed random seed.

\subsection*{Preprocessing and sentence representation}

All experiments were implemented in Python using \texttt{stanza}~\cite{qi-etal-2020-stanza}:

\begin{itemize}
    \item Articles were segmented into sentences with the French \texttt{stanza} pipeline and each sentence was assigned an \texttt{article\_id} and a sentence index (\texttt{sent\_id}).
    \item We constructed local context windows by sorting sentences by \texttt{article\_id, sent\_id} and, for each sentence, concatenating up to \emph{three} preceding and \emph{three} following sentences from the same article into a single \texttt{context\_text}.
    \item In \texttt{context\_text}, the target sentence was wrapped in \texttt{<tgt> ... </tgt>} and neighboring sentences were separated by the delimiter \texttt{" </s> "}.
\end{itemize}

\subsection*{LLM pseudo-labeling with Qwen}

All Qwen3-30B calls were run locally on local GPU servers (no prompts or texts were sent to external APIs). The exact system and user prompts used to query Qwen for binary and six-way stance decisions are released with the code at \url{https://gitlab.idiap.ch/socialcomputing/politics-of-questions}.

\paragraph{Candidate detection.} To reduce LLM calls, we first flagged \emph{candidate} interrogative sentences with a high-recall heuristic:

\begin{itemize}
    \item Presence of a question mark \enquote{?}.
    \item Sentence-initial interrogative patterns (case-insensitive), e.g. \emph{comment}, \emph{pourquoi}, \emph{combien}, \emph{est-ce que}, \emph{y a-t-il}, \emph{faut-il}, \emph{peut-on}, \emph{doit-on}, \emph{que faire}, \emph{que va-t-il}.
    \item Question-raising verb patterns, e.g. \emph{on peut se demander}, \emph{on est en droit de se demander}, \emph{(se) demande(r)}, \emph{(s') interroger}.
    \item Question-raising noun patterns, e.g. \emph{pose la question}, \emph{soulève la question}, \emph{remet en question}, \emph{la question se pose/reste/demeure}, \emph{question en suspens}, \emph{questions sans réponse}, \emph{reste à savoir}.
\end{itemize}

In addition, a random subsample of non-candidate sentences (25\% as many as candidates) was sent to Qwen for calibration.

\paragraph{Binary interrogative stance (teacher).}

For the binary decision (\emph{interrogative stance} vs.\ \emph{non-interrogative}), we used:

\begin{itemize}
    \item Model: \texttt{Qwen/Qwen3-30B-A3B-Instruct-2507}.
    \item Input: a system prompt defining \enquote{interrogative stance} and a user prompt with up to three sentences of left context, the target sentence marked as \enquote{Phrase cible}, and up to three sentences of right context.
    \item Decoding: greedy (\texttt{temperature=0.0}, \texttt{top\_p=1.0}), \texttt{max\_new\_tokens=64}, batch size 8.
    \item Output: JSON with \texttt{"is\_interrogative"} (boolean) and \texttt{"confidence"} $\in \{0.2, 0.5, 0.8, 0.95\}$.
\end{itemize}

\paragraph{Six-way stance classification (teacher).}

For sentences judged interrogative by the binary teacher, we used the same Qwen model with a different system prompt to assign one of six stance labels:

\begin{itemize}
    \item Labels: \texttt{"information-seeking"}, \texttt{"rhetorical"}, \texttt{"leading"}, \texttt{"tag"}, \texttt{"echo-clarification"}, \texttt{"framing-procedural"}.
    \item Input: same context format as above; output: JSON with \texttt{"label"} in the six-way set and \texttt{"confidence"} $\in \{0.2, 0.5, 0.8, 0.95\}$.
\end{itemize}

For training student models, we retained only high-confidence LLM labels:

\begin{itemize}
    \item Binary: positives (\texttt{True}) and negatives (\texttt{False}) with confidence $\geq 0.7$.
    \item Stance: six-way labels with confidence $\geq 0.7$.
\end{itemize}

\subsection*{Student models: CamemBERT classifiers}

\paragraph{Binary interrogative detector.}

The binary student model is a CamemBERT-large sequence classifier:

\begin{itemize}
    \item Base encoder: \texttt{camembert/camembert-large}.
    \item Labels: \{\texttt{"non-interrogative"}, \texttt{"interrogative"}\}.
    \item Training data: high-confidence Qwen binary labels, excluding all articles in the human evaluation sets.
    \item Input: \texttt{context\_text} with a $\pm 3$-sentence window and \texttt{<tgt>} markers, max sequence length 512 tokens.
    \item Split: Split on \texttt{article\_id} with 10\% held out for validation (no article overlap).
    \item Optimization: AdamW, learning rate $2 \times 10^{-5}$, weight decay 0.01, warmup ratio 0.06, batch sizes 16 (train) and 32 (validation), up to 20 epochs with early stopping (patience 3) on validation macro-F1. Weighted cross-entropy loss to balance classes.
\end{itemize}

\paragraph{Six-way stance classifier.}

The stance student model uses the same base encoder and input representation:

\begin{itemize}
    \item Labels: the six stance types above.
    \item Training data: high-confidence Qwen stance labels (confidence $\geq 0.7$) for sentences judged interrogative by the binary teacher, again excluding evaluation articles.
    \item Optimization: same schedule as the binary model, with class weights computed per stance label to compensate for imbalance (leading and tag are rare).
\end{itemize}

Both models were later used for two-step inference on the full sentence-level corpus, with a binary confidence threshold of 0.7 to gate sentences into the six-way stance classifier.

\subsection*{Answer span identification}

Answer spans were identified with a simple embedding-based search using CamemBERT-large:

\begin{itemize}
    \item Model: \texttt{camembert/camembert-large}.
    \item Embedding input: for each sentence, an \texttt{embed\_text} context consisting of the target sentence with a $\pm 1$-sentence window (same-article neighbors), with the target wrapped in \texttt{<tgt> ... </tgt>} and sentences separated by \texttt{" </s> "}.
    \item Tokenization: truncation to 256 tokens, padding to batch max length, batch size 64.
    \item Sentence embedding: mean pooling of the last hidden layer over non-padding tokens, followed by $\ell_2$ normalization.
\end{itemize}

Within each article, we:

\begin{itemize}
    \item Treated as \emph{questions of interest} all sentences whose stance label was one of the six interrogative types with stance confidence $\geq 0.7$.
    \item Grouped consecutive question sentences (no gap in \texttt{sent\_id}) into local question groups.
\end{itemize}

For each question group within an article:

\begin{enumerate}
    \item Compute a group embedding as the average of the normalized sentence embeddings in the group, re-normalized to unit length.
    \item Precompute cumulative sums over the article’s sentence embeddings to allow fast average embeddings over any contiguous window.
    \item Search only among subsequent sentences up to 15 sentences ahead of the last question sentence in the group.
    \item For each candidate window length $L \in \{1, 2, 3, 4, 5\}$ and each possible start position, compute the mean embedding and its cosine similarity with the group embedding.
    \item If the best window has cosine similarity $\geq 0.40$, treat it as the answer span. Sensitivity checks reported in Appendix~\ref{app:quant} show that the stored similarity scores are sharply bimodal, so the main answerability estimates are effectively invariant across a broad range of thresholds. Otherwise mark the group as unanswered.
\end{enumerate}

For each interrogative sentence we stored whether an answer was found, the similarity score, and the sentence indices and concatenated text of the answer span. These values underlie the article-level answerability and dialogicity indices.

\subsection*{BERTopic configuration}

Article-level topics were derived with BERTopic on CamemBERT-large article embeddings, using GPU-accelerated UMAP and HDBSCAN:

\begin{itemize}
    \item \textbf{model} $\sim$1M articles: UMAP \texttt{n\_components=5}, \texttt{n\_neighbors=50}, \texttt{min\_dist=0.0}, \texttt{metric="cosine"}; HDBSCAN \texttt{min\_cluster\_size=200}, \texttt{min\_samples=20}.
\end{itemize}

For each article we recorded its topic ID and the maximum topic probability. Frequently occurring topics were manually grouped into eight meta-topics used in the main analysis.
For transparency, the 100 most frequent BERTopic clusters were manually assigned to the eight meta-topics by inspecting each cluster’s top keywords and representative articles. Clusters with mixed semantics were assigned according to their dominant content after joint author inspection.

\subsection*{Named-entity recognition with GLiNER}

We applied NER to questions and their detected answer spans using the multilingual GLiNER model:

\begin{itemize}
    \item Model: \texttt{urchade/gliner\_multi-v2.1}.
    \item Label set (coarse and robust): \textsc{person}, \textsc{organization}, \textsc{location}, \textsc{nationality or religious or political group}, \textsc{generic social group}, \textsc{public or audience}, \textsc{event}.
    \item Inference settings: batch size 16 and score threshold 0.5.
\end{itemize}

NER was run on the QA-enhanced sentence files produced by the answer-identification step. We proceeded as follows:

\begin{itemize}
    \item \textbf{Selecting questions.} We selected as \emph{questions of interest} the same sentences used in the answer-identification step: those whose CamemBERT stance label was one of the six interrogative types and whose stance confidence was at least 0.7.
    \item \textbf{Question-side NER.} For each selected question, we built a short context string:
    \begin{quote}
    previous sentence (if any) + question sentence + next sentence (if any),
    \end{quote}
    \item \textbf{Answer-side NER.} For questions marked as \texttt{has\_answer = True}, we passed the concatenated \texttt{answer\_span\_text} (without extra context) to GLiNER.
    \item We stored the resulting entities in columns as lists of entity records.
\end{itemize}

\section{Additional quantitative summaries}
\label{app:quant}

Table~\ref{tab:meta_topics} provides a detailed overview of meta-topics, showing how interrogative density and the presence of organizations, locations, and events vary across article domains. Table~\ref{tab:answ_dialog} summarizes answerability and dialogicity at the stance level, including the overall shares of unanswered questions and of questions answered internally versus via quoted speech.

\begin{table*}[h]
    \centering
    \rowcolors{2}{}{gray!15}
    \begin{tabular}{lrrrr}
        \toprule
        Meta-topic & Articles & Mean interrog. index & \% Q with ORG & \% with LOC/EVENT \\
        \midrule
        Local news                                 &  39{,}662 & 0.0298 & 38.3 & 36.3 / 28.4 \\
        Professional sports                        & 181{,}056 & 0.0290 & 51.8 & 33.4 / 33.4 \\
        Lifestyle, entertainment \& people         &  44{,}533 & 0.0271 & 27.8 & 25.8 / 26.1 \\
        \emph{Faits divers}                        &  29{,}932 & 0.0220 & 37.7 & 39.6 / 21.5 \\
        National / local politics                  & 118{,}167 & 0.0214 & 47.1 & 35.7 / 23.8 \\
        Technology                                 &  12{,}104 & 0.0184 & 44.7 & 17.7 / 20.6 \\
        Business \& economy                        &  25{,}412 & 0.0182 & 57.9 & 29.7 / 24.8 \\
        Geopolitics                                &  94{,}190 & 0.0180 & 42.9 & 46.7 / 31.0 \\
        \bottomrule
    \end{tabular}
    \caption{Meta-topic overview: article volume, interrogative density, and institutional, geographic, and event anchoring of questions (Q).}
    \label{tab:meta_topics}
\end{table*}

\begin{table}[h]
    \centering
    \rowcolors{2}{}{gray!15}
    \begin{tabular}{lrr}
        \toprule
        Stance & $N_Q$ & \% answered \\
        \midrule
        information-seeking   & 153{,}404 & 97.2 \\
        echo-clarification    &  70{,}281 & 97.4 \\
        rhetorical            & 107{,}862 & 95.5 \\
        leading               &  16{,}221 & 95.4 \\
        tag                   &  16{,}116 & 95.1 \\
        framing-procedural    & 396{,}298 & 94.8 \\
        \midrule
        \rowcolor{white}
        All stances           & 760{,}182 & 95.6 \\
        \bottomrule
    \end{tabular}

    \vspace{1em}
    \rowcolors{2}{}{gray!15}
    \begin{tabular}{lrr}
        \toprule
        Category & $N_Q$ & \% of interrogatives \\
        \midrule
        Unanswered            &  33{,}271 & 4.4 \\
        Answered (internal)   & 610{,}209 & 80.3 \\
        Answered (via quotes) & 116{,}702 & 15.4 \\
        \bottomrule
    \end{tabular}
    
    \caption{Answerability and dialogicity of interrogative stances.}
    \label{tab:answ_dialog}
\end{table}

\subsection*{Sensitivity to confidence and answer-similarity thresholds}

To assess the robustness of the main corpus-level findings to threshold choices, we conducted two simple sensitivity checks. First, we varied the confidence threshold used to retain interrogative predictions from the binary and six-way stance classifiers. As shown in Table~\ref{tab:confidence_sensitivity}, stricter thresholds reduce the absolute number of detected interrogatives, as expected, but the overall prevalence estimates remain in the same range: interrogatives remain sparse, with mean article-level interrogative indices between 0.024 and 0.026.

\begin{table}[h]
    \centering
    \rowcolors{2}{}{gray!15}
    \begin{tabular}{lrrr}
        \toprule
        Confidence & $N_Q$ & \% of sentences & Mean $ID_a$ \\
        \midrule
        0.6 & 785{,}014 & 3.34 & 0.0261 \\
        \textbf{0.7} & 760{,}182 & 3.23 & 0.0252 \\
        0.8 & 714{,}023 & 3.03 & 0.0236 \\
        \bottomrule
    \end{tabular}
    
    \caption{Sensitivity of interrogative prevalence estimates to the confidence threshold used for the interrogative predictions.}
    \label{tab:confidence_sensitivity}
\end{table}

Second, we varied the cosine-similarity threshold used in answer matching. The stored similarity scores were sharply separated rather than clustered around the baseline cutoff: all unmatched cases fell below 0.35, whereas matched spans were overwhelmingly assigned much higher scores. Consequently, answerability estimates were identical for thresholds from 0.05 to 0.80 and declined only marginally at stricter cutoffs (Table~\ref{tab:answer_sensitivity}). This indicates that the high answerability rate is not driven by a narrow threshold choice. At the same time, because the answer-matching procedure captures semantic relatedness rather than manually verified resolution, these values should still be interpreted as heuristic proxies for textual uptake rather than literal estimates of fully resolved answerhood.

\begin{table}[h]
    \centering
    \rowcolors{2}{}{gray!15}
    \begin{tabular}{lrrr}
        \toprule
        Similarity & answered & internal & via quotes \\
        \midrule
        0.05  & 95.6\% & 80.3\% & 15.4\% \\
        \textbf{0.40}  &  95.6\% & 80.3\% & 15.4\% \\
        0.80  & 95.6\%  & 80.3\% & 15.4\% \\
        0.95  & 94.9\%  & 79.6\% & 15.3\% \\
        0.975 & 82.5\%  & 69.1\% & 13.5\% \\
        \bottomrule
    \end{tabular}
    \caption{Sensitivity of answerability and dialogicity estimates to the cosine-similarity threshold used for answer matching.}
    \label{tab:answer_sensitivity}
\end{table}

\subsection*{Manual spot check of answer spans}

To complement the threshold sensitivity analysis, we manually audited 50 randomly sampled question groups from 50 different articles, including 40 predicted answered and 10 predicted unanswered cases split across the local and national corpora. Table~\ref{tab:answer_manual_spotcheck} summarizes the results.

\begin{table}[h]
    \centering
    \small

    \rowcolors{2}{}{gray!15}
    \begin{tabular}{lccc}
        \toprule
        Pred. answered & Local & National & Total \\
        \midrule
        Clear answer                  & 11 (55\%)   & 14 (70\%)   & 25 (62.5\%) \\
        Partial answer                &  5 (25\%)   &  4 (20\%)   &  9 (22.5\%) \\
        Answer elsewhere   &  1 (5\%)    &  2 (10\%)   &  3 (7.5\%)  \\
        No genuine answer             &  3 (15\%)   &  0 (0\%)    &  3 (7.5\%)  \\
        \bottomrule
    \end{tabular}

    \vspace{1em}

    \rowcolors{2}{}{gray!15}
    \begin{tabular}{lccc}
        \toprule
        Pred. unanswered & Local & National & Total \\
        \midrule
        Correct. unanswered & 5 (100\%) & 5 (100\%) & 10 (100\%) \\
        Missed answer & 0 (0\%)   & 0 (0\%)   & 0 (0\%)    \\
        \bottomrule
    \end{tabular}
    \caption{Manual spot check of the answer heuristic on 50 randomly sampled questions from 50 different articles.}
    \label{tab:answer_manual_spotcheck}
\end{table}

Overall, 34 of the 40 predicted answered cases (85.0\%) corresponded to clear or partial answers, and in 37 of 40 cases (92.5\%) the article contained an answer somewhere, while all 10 predicted unanswered cases were confirmed as genuinely unanswered.

\section{Model and annotation quality}
\label{app:eval}

Table~\ref{tab:model_iaa} complements the main-text summary of the performance assessment of the classification pipeline and the inter-annotator agreement.

\begin{table}[h]
    \centering
    \rowcolors{2}{}{gray!15}
    \begin{tabular}{lr}
        \toprule
        \multicolumn{2}{c}{\textbf{Binary interrogative detector}} \\
        \midrule
        \rowcolor{white}
        Metric & Value  \\
        \midrule
        Evaluation sentences          & 8{,}399  \\
        Accuracy                      & 0.97     \\
        Precision (interrogative)     & 0.76     \\
        Recall (interrogative)        & 0.80     \\
        F1 (interrogative)            & 0.78     \\
        \midrule
        \rowcolor{white}
        \multicolumn{2}{c}{\textbf{Six-way stance classifier}} \\
        \midrule
        Metric & Value  \\
        \midrule
        Evaluation interrogatives     & 516      \\
        Macro-F1                      & 0.51     \\
        Micro-F1                      & 0.51     \\
        \midrule
        \multicolumn{2}{c}{\textbf{Inter-annotator agreement (stance)}} \\
        \midrule
        \rowcolor{white}
        Metric & Value  \\
        \midrule
        Double-coded articles         & 98       \\
        Matched interrogative units   & 204      \\
        Jaccard overlap (spans)       & 0.83     \\
        Accuracy (stance labels)      & 0.84     \\
        Cohen’s $\kappa$              & 0.78     \\
        \bottomrule
    \end{tabular}
    \caption{Summary of model performance and inter-annotator agreement.}
    \label{tab:model_iaa}
\end{table}

\begin{figure}[h]
    \centering
    \includegraphics[width=\linewidth]{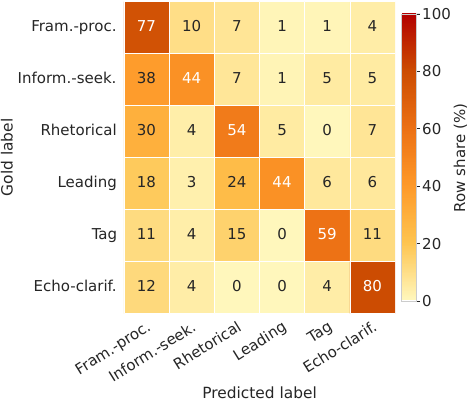}
    \caption{Row-normalized conditional confusion matrix for the six-way stance classifier.}
    \label{fig:stance_confusion_matrix}
\end{figure}

Table~\ref{tab:stance_per_class} reports per-class precision, recall, F1, and support for the six-way stance classifier on the full gold interrogative evaluation set ($n=516$). Figure~\ref{fig:stance_confusion_matrix} provides a complementary conditional confusion matrix computed only on gold interrogative sentences that were passed to the six-way stage by the binary detector. Because binary false negatives are excluded from this conditional view, the row-normalized diagonal values in Figure~\ref{fig:stance_confusion_matrix} should not be read as recalls and are therefore not directly comparable to the recall values reported in Table~\ref{tab:stance_per_class}.

\begin{table}[h]
    \centering

    \rowcolors{2}{}{gray!15}
    \begin{tabular}{lrrrr}
        \toprule
        Stance & Precision & Recall & F1 & Support \\
        \midrule
        Fram.-proc. & 0.45 & 0.59 & 0.51 & 129 \\
        Inform.-seek. & 0.80 & 0.37 & 0.51 & 171 \\
        Rhetorical & 0.60 & 0.39 & 0.47 & 113 \\
        Leading & 0.68 & 0.38 & 0.49 & 39 \\
        Tag & 0.59 & 0.50 & 0.54 & 32 \\
        Echo-clarif. & 0.48 & 0.62 & 0.54 & 32 \\
        \bottomrule
    \end{tabular}
    \caption{Per-class performance of the six-way stance classifier on the gold-standard evaluation set.}
    \label{tab:stance_per_class}
\end{table}

The conditional confusion matrix shows that most residual errors arise among pragmatically adjacent stance types. Information-seeking and rhetorical questions are often redistributed toward framing-procedural once a sentence has been recognized as interrogative, suggesting that framing-procedural partly functions as a broad residual category at the six-way stage. Leading questions are likewise frequently mapped to rhetorical or framing-procedural, consistent with the qualitative proximity between strongly oriented evaluation and more overtly leading design. By contrast, echo-clarification remains comparatively distinctive, while tag questions are moderately recovered but rely on a relatively small number of gold cases. More broadly, these disagreements should be interpreted in light of the intrinsic ambiguity of the task: many interrogatives plausibly support more than one stance reading, and post-hoc inspection of mismatches suggests that some model-gold discrepancies reflect borderline or multifunctional cases rather than unequivocal classification failures. We therefore use the six-way predictions as approximate corpus-level indicators and read fine-grained contrasts among adjacent stance types with appropriate caution.

\section{Additional qualitative observations and illustrative examples}
\label{app:qual_quotes}

For reasons of space, the main text can only briefly refer to individual examples from the qualitative subcorpus. This appendix presents a small set of additional excerpts that were cut from the main text but are analytically representative of broader patterns. Each example is accompanied by a short comment on why it is interesting in terms of interrogative stance, macro-function, and the indices introduced in Section~\ref{sec:method}.

\subsection*{Framing local decline as a problem to be solved}

Coverage of local socioeconomic trends sometimes uses interrogatives to recast descriptive statistics as collective problems. In a Valaisan article on the decline in apprenticeships, the lead introduces a factual decrease and then asks, in translation:

\begin{quote}
\emph{\enquote{Ten percent fewer apprentices: how can Valais fix this?}}
\end{quote}

The question is classified as \emph{framing-procedural} but it also carries an evaluative presupposition: the decline is implicitly treated as undesirable and in need of remedy. It illustrates how factual presuppositions (\emph{there has been a 10\% drop}) can be combined with verbs like \enquote{fix} or \enquote{remedy} to move from neutral description to engaged framing, while still remaining relatively low in face-threat and open-ended in terms of specific solutions.

\subsection*{Engaged framing and mild conduciveness in hyper-local politics}

Hyper-local outlets sometimes use more pointed interrogatives to highlight institutional strain. In a feature on parliamentary workload in a Swiss canton, a journalist reports complaints about the number of motions and then asks:

\begin{quote}
\emph{\enquote{Are Valais deputies suffocating their own Parliament?}} \\
\emph{\enquote{Is the parliamentary machine at risk of overload?}}
\end{quote}

These polar questions are annotated as \emph{leading} and mapped to \emph{Framing/agenda-setting}. They embed evaluative metaphors (\enquote{suffocating}, \enquote{overload}) and present a narrow choice between a problematic and a non-problematic reading of the situation. In our indices, such questions contribute to higher interrogative density and a slightly higher share of leading stances in hyper-local political coverage, while remaining substantially less confrontational than the adversarial formats documented in some broadcast traditions.

\subsection*{Authority positioning in expert interviews}

In interviews with experts, interrogatives often enact an epistemic asymmetry in which the journalist adopts a knowledge-seeking role. In a piece on the Russian Orthodox Church’s participation in ecumenical dialogue, after summarizing calls to sanction the Moscow Patriarchate, the interviewer turns to a scholar and asks:

\begin{quote}
\emph{\enquote{If the Russian Orthodox Church does not really want to take part in ecumenical debate, what interest does it have in participating at all?}}
\end{quote}

This is labeled \emph{information-seeking} and mapped to the macro-axis of \emph{Authority positioning}. The journalist’s epistemic deficit is made explicit through the \enquote{what interest} formulation, while the presupposition about reluctance is attributed to prior sources rather than asserted in the journalist’s own voice. The answer appears as an extended quoted response, contributing to high external dialogicity and illustrating the \enquote{epistemic extraction} mode in which questions are used to elicit explanations rather than to confront.

\subsection*{Personalizing cross-border relations}

Some analytical pieces use interrogatives to personalize complex international relationships around a single political figure. In a Canadian analysis of U.S.–Canada relations under the prospect of a second Trump presidency, the closing paragraph asks:

\begin{quote}
\emph{\enquote{Did Donald Trump and his advisers leave Washington with a better understanding of Canada’s position?}}
\end{quote}

Throughout the article, Trump is repeatedly named and foregrounded, and this final question recasts a structural diplomatic issue in terms of what one individual has or has not learned. In our terminology, the interrogative is \emph{framing-procedural} and contributes to a highly personalized interrogative space: a complex bilateral relationship is summarized as a question about a single actor’s knowledge state.

\subsection*{Localization and collectivization of policy debates}

Interrogatives can also localize national policy debates and link them explicitly to a specific public. In Quebec coverage of federal climate policy, after describing a proposed abolition of a federal levy, a journalist asks:

\begin{quote}
\emph{\enquote{Would abolishing this federal contribution cause Quebec’s carbon market to collapse?}}
\end{quote}

This information-seeking question is also a framing move: it connects an abstract fiscal change to potential local consequences and names a specific place and collective concern (the provincial carbon market). In our NER-based indices, such cases combine \textsc{location} and \textsc{public/audience} mentions, illustrating the relatively small but significant share of interrogatives that directly implicate broad publics rather than only elites or institutions.

\subsection*{Everyday dialogicity in hyper-local features}

Finally, some hyper-local feature stories include conversational question–answer pairs that foreground everyday interactions among non-elite actors. In a Swiss piece about a sewing class, the journalist recounts an exchange between a novice participant and his instructor:

\begin{quote}
\emph{\enquote{Can I use my own sewing machine?} he asks.} \\
\emph{\enquote{Of course, as long as it is in good working order,} the teacher replies.}
\end{quote}

These are straightforward \emph{information-seeking} interrogatives realized in direct speech. They contribute to external dialogicity and to the presence of non-elite \textsc{person} entities in both questions and answers. While numerically marginal in the corpus, such examples highlight how interrogatives can be used to stage ordinary voices and lived experience, especially in hyper-local reporting.

Taken together, these additional excerpts show how fine-grained qualitative distinctions, between neutral and engaged framing, epistemic extraction and narrative authority, personalization and localization, underlie the aggregate patterns reported in Section~\ref{sec:results}.

\end{document}